\documentclass{article}
\usepackage{conference,times}
\usepackage[T1]{fontenc}
\usepackage{times}

\usepackage{amsmath,amsfonts,bm}

\def\eqref#1{equation~\ref{#1}}
\def\1{\bm{1}}

\def\va{{\bm{a}}}
\def\vb{{\bm{b}}}
\def\vc{{\bm{c}}}

\def\ve{{\bm{e}}}

\def\vg{{\bm{g}}}

\def\vt{{\bm{t}}}

\def\vx{{\bm{x}}}

\DeclareMathAlphabet{\mathsfit}{\encodingdefault}{\sfdefault}{m}{sl}
\SetMathAlphabet{\mathsfit}{bold}{\encodingdefault}{\sfdefault}{bx}{n}

\renewcommand{\eqref}[1]{(\ref{#1})}

\usepackage{url}
\usepackage{caption}
\usepackage{graphicx}
\usepackage{enumitem}
\usepackage{booktabs}
\usepackage{float}
\usepackage{hyperref}
\usepackage{amssymb}
\newcommand{\best}[1]{\ensuremath{\mathbf{#1}}}
\newcommand{\vomega}{\bm{\omega}}   % Euler 角，替代 \vr
\newcommand{\veta}{\bm{\eta}}       % action mask，替代 \mathcal{M}_t
\title{Organize Primitives into Semantic Parts: Reinforcement Reasoning for 3D Segmentation}

\author{
{\bfseries\begin{tabular}[t]{@{}l@{\hspace{0.75em}}l@{\hspace{0.75em}}l@{\hspace{0.75em}}l@{}}
Xiaoming Gong$^{1}$\thanks{Equal contribution.} & Ruoyu Wu$^{2}$\footnotemark[1] & Zhenhong Sun$^{3}$ & Chunlin Chen$^{1}$\\
Daoyi Dong$^{4}$ & Huadong Mo$^{2}$ & Zhi Wang$^{1}$ & Hongdong Li$^{3}$
\end{tabular}}\\[0.6ex]
{\normalfont\small $^{1}$Nanjing University, China \quad $^{2}$University of New South Wales, Canberra, Australia}\\
{\normalfont\small $^{3}$Australian National University, Australia}\\
{\normalfont\small $^{4}$University of Technology Sydney, Australia}\\[0.4ex]
}
\date{}

\iclrfinalcopy
\begin{document}

\maketitle
\lhead{}

\begin{abstract}
Primitive-based 3D segmentation offers a compact and explicit alternative to dense surface prediction, naturally supporting structural abstraction and boundary localization.
However, geometric decomposition alone does not determine how primitives should be organized into semantic parts: a single part may span multiple primitives, while geometrically similar or touching primitives may belong to different parts.
We therefore introduce \textbf{RePart} (\textbf{Re}inforcement \textbf{Part} Reasoning), which formulates primitive-to-part organization as a finite-horizon Markov decision process and learns semantic organization through trajectory-level reinforcement reasoning.
RePart constructs a \emph{Composable Primitive Workspace} from fine-grained superquadrics and applies a merge-and-stop policy whose decisions are optimized by their downstream effects on the resulting partition rather than local primitive compatibility.
The inferred part identities are then mapped back to the original mesh through \emph{Boundary-Aware Surface Labeling}, preserving accurate surface boundaries beyond the primitive approximation.
On PartNet, RePart achieves the strongest results across all four aggregate partition metrics; on 3DCoMPaT++, it obtains the highest RI and SC without target-dataset fine-tuning.
These results demonstrate that reinforcement reasoning provides an effective mechanism for organizing geometric primitives into semantic parts while retaining dense segmentation accuracy.
Code is available at \href{https://github.com/EngineeringAI-LAB/RePart}{https://github.com/EngineeringAI-LAB/RePart}.
\end{abstract}

\section{Introduction}
3D part segmentation has advanced rapidly through complementary paradigms, including geometric primitive decomposition
\citep{paschalidou2021neural,wu2022primitive,monnier2023differentiable,liu2023deformer,liu2023marching},
learned 3D representations
\citep{ma2025find,liu2025partfield,hadgi2026patchalign3d},
and semantic transfer from vision--language and multi-view foundation models
\citep{liu2023partslip,umam2024partdistill,yang2024sampart3d,zhu2026partsam}.
These advances have improved the flexibility and generalization of 3D part understanding.
Meanwhile, LLM-driven 3D agents and interactive systems \citep{qi2024shapellm} are broadening the role of segmentation: predicted parts increasingly serve not only as dense outputs, but also as entities for reasoning, selection, editing \citep{ganeshan2024parsel}, and manipulation \citep{geng2023gapartnet}.
This shift demands representations that expose compact object structure for high-level reasoning while remaining grounded in the original geometry.
Geometric primitives are well suited to these requirements, replacing dense surface elements with spatially coherent units that encode shape, scale, pose, orientation, and volumetric extent.

Primitive-based 3D decomposition has evolved from learned and deformable shape abstractions
\citep{paschalidou2021neural,monnier2023differentiable,liu2023deformer},
to explicit primitive fitting
\citep{wu2022primitive,liu2023marching},
and structure-aware abstraction
\citep{wang2025light,fedele2025superdec}.
These methods naturally support \emph{Structural Abstraction}: dense geometry is represented by compact explicit components at a tractable granularity \citep{paschalidou2019superquadrics}.
They also support \emph{Boundary Localization}, since each primitive retains spatial support that maps back to the surface \citep{li2019supervised}.
Yet, as parts increasingly serve as meaningful entities rather than dense outputs, a third requirement becomes critical: \emph{Semantic Organization}.
Primitive decomposition is mainly governed by geometric fit and structural regularity, which do not determine how primitives compose into semantic parts.
A single part may span multiple primitives, while nearby primitives with similar geometry may belong to different parts.
Thus, the key unresolved problem is how to organize primitives into semantic parts \citep{luo2020learning}.

\emph{Semantic Organization} cannot be reduced to independent or locally optimal grouping decisions.
Pairwise classification assumes that mergeability can be judged in isolation, while greedy agglomeration optimizes only immediate gains and cannot recover from early irreversible mistakes.
Supervised sequence learning is also ill-posed, because ground-truth segmentation specifies the desired final partition but not a unique merge trajectory.
Starting from an over-segmented primitive partition, each merge changes group geometry, neighborhood relations, and the candidate actions available thereafter, so the value of an early decision is ultimately determined by the partition it enables.
We therefore formulate semantic organization as a finite-horizon Markov decision process \citep{puterman1990markov}.
This formulation makes reinforcement learning (RL) well matched to the problem: it assigns credit from the quality of the resulting partition back through the sequence of merge-and-stop decisions \citep{sutton1998reinforcement}, rather than requiring local pair labels or a prescribed reasoning path.

Building on this formulation, we propose \textbf{RePart} (\textbf{Re}inforcement \textbf{Part} Reasoning), which places \emph{Semantic Organization} at the center of primitive-based 3D segmentation and adapts both the reasoning representation and dense output around it.
{Composable Primitive Workspace} realizes \emph{Structural Abstraction} by constructing a fine-grained SQ workspace with Marching-Primitives, then fixing the primitive geometry and surface correspondence to provide stable, composable units for reasoning without committing to the final part partition.
{Reinforcement-Learning Part Reasoning} realizes \emph{Semantic Organization} by casting SQ grouping as a merge-and-stop decision process and optimizing the policy with PPO, allowing each merge to be evaluated by its downstream effect on the evolving and final partition rather than local compatibility alone.
{Boundary-Aware Surface Labeling} realizes \emph{Boundary Localization} by transferring the inferred part identities to curvature-consistent mesh regions using grouped-SQ surface proximity and volumetric evidence, so that the final segmentation follows the original surface geometry instead of the approximate primitive boundaries.

The contributions of this paper are summarized as follows:
\begin{itemize}[leftmargin=*, noitemsep, nolistsep]
    \item We formulate primitive-to-part organization as a finite-horizon Markov decision process, enabling semantic grouping without requiring predefined merge sequences.
    
    \item We introduce \textbf{Reinforcement-Learning Part Reasoning} to optimize merge-and-stop decisions through trajectory-level credit assignment rather than local primitive compatibility.
    
    \item We couple this reasoning with a \textbf{Composable Primitive Workspace} and \textbf{Boundary-Aware Surface Labeling}, bridging compact primitive reasoning and precise mesh segmentation.
\end{itemize}

\begin{figure}[t]
    \centering
     \includegraphics[width=\linewidth]{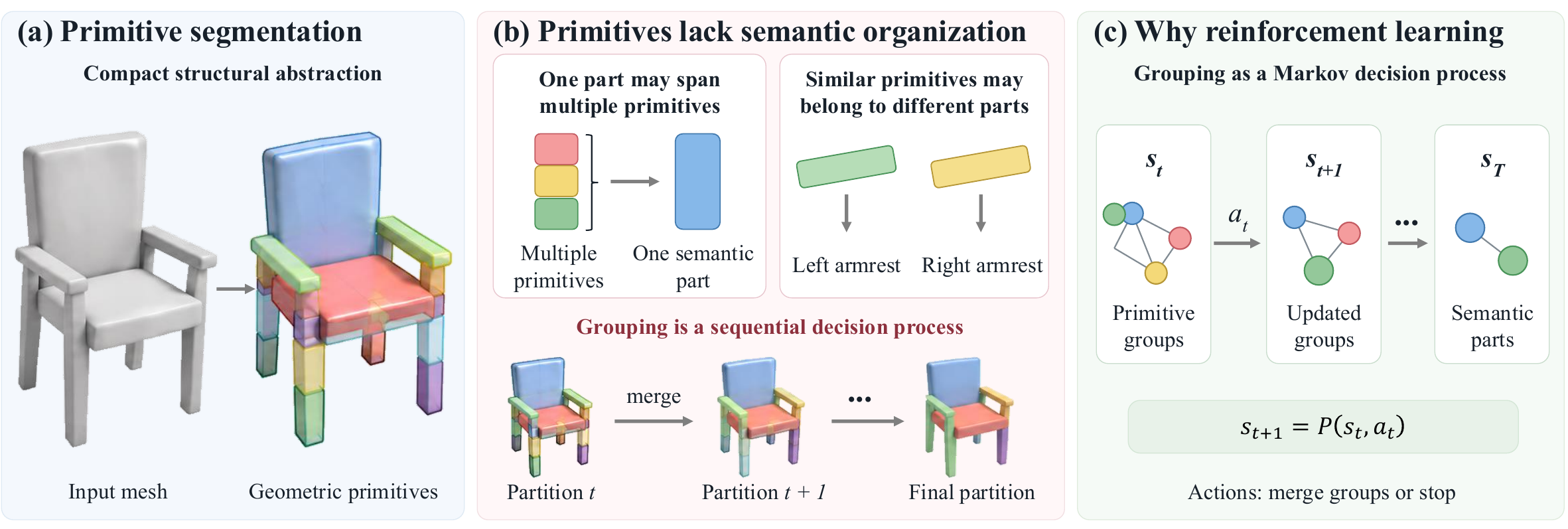}
     \caption{Primitive segmentation exposes compact structural abstractions, while organizing primitives into semantic parts requires sequential decisions, making reinforcement learning a natural fit.}
    \label{fig:RePart-concept}
\end{figure}

\section{Related Work}

\textbf{3D Part Segmentation.}
Recent 3D part segmentation increasingly targets low-shot, class-agnostic, and open-world settings. Vision--language and multi-view foundation models transfer semantic cues to 3D \citep{liu2023partslip,umam2024partdistill,yang2024sampart3d,tang2024segment}, while large-scale 3D supervision and learned representations improve generalization and controllable granularity \citep{ma2025find,liu2025partfield,zhu2026partsam,hadgi2026patchalign3d,su2026s,deng2026geosam2}. These methods primarily predict dense part assignments. RePart instead reasons over compact geometric primitives, separating semantic organization from surface labeling.

\textbf{Primitive-Based 3D Abstraction.}
Primitive abstraction represents 3D geometry with compact, explicit components, ranging from classical superquadric recovery \citep{leonardis1997superquadrics} to learned, probabilistic, deformable, and differentiable formulations \citep{paschalidou2021neural,wu2022primitive,liu2023deformer,monnier2023differentiable}. Later work models hierarchical structure or improves geometry-aware decomposition \citep{niu2022rim,ruan2025learning,liu2025wir3d}, while MPS, Light-SQ, and SUPERDEC further develop SQ-based local abstraction and structural organization \citep{liu2023marching,wang2025light,fedele2025superdec}. GaussianGPT uses explicit Gaussian primitives as discrete tokens for scene generation \citep{von2026gaussiangpt}. RePart instead treats primitives as composable reasoning units and focuses on organizing them into semantic parts.

\textbf{Sequential Primitive Organization.}
Classical mesh segmentation and hierarchical decomposition progressively merge geometric regions or components \citep{attene2006hierarchical,niu2022rim,du2024generative,ye2025primitiveanything}. Learning to Group formulates bottom-up grouping as a contextual bandit with local mergeability rewards \citep{luo2020learning,brehmer2020hierarchical}, while reinforcement learning has also been explored for hierarchical clustering and 3D reconstruction \citep{brehmer2020hierarchical,li20233d}. RePart differs by formulating primitive-to-part organization as a finite-horizon merge-and-stop Markov decision process, where irreversible decisions are optimized through their downstream effect on the resulting partition.

\section{Method}

\subsection{Problem Formulation and Overview}

Given an input triangular mesh $\mathcal{O}=(\mathcal{V},\mathcal{F})$, our goal is to predict a class-agnostic partition
\[
\hat{Y}^{\mathcal{F}}=\{\hat{y}^{\mathcal{F}}_f\}_{f\in\mathcal{F}}
\]
of its faces into an unknown number of parts; its numeric labels are defined only up to permutation. During training, we additionally sample a point set
$X=\{\vx_i\}_{i=1}^{N}$ from the mesh surface and associate each point with a ground-truth part label
$Y=\{y_i\}_{i=1}^{N}$.

Primitive-based part segmentation exposes three coupled gaps across representation, organization, and boundary localization: dense mesh elements preserve surface detail but are unnecessarily fine-grained for global structural reasoning; fitted primitives provide compact geometric units but do not determine part identities; and primitive groups do not directly localize accurate mesh boundaries. RePart addresses these gaps through a \emph{Composable Primitive Workspace} for structural abstraction, \emph{Reinforcement-Learning Part Reasoning} for semantic organization, and \emph{Boundary-Aware Surface Labeling} for boundary localization:

\[
\mathcal{O}
\;\xrightarrow{\substack{\text{Composable Primitive}\\\text{Workspace}}}\;
(\mathcal{W},\Pi_0)
\;\xrightarrow{\substack{\text{RL Part}\\\text{Reasoning}}}\;
\Pi_T
\;\xrightarrow{\substack{\text{Boundary-Aware}\\\text{Surface Labeling}}}\;
\hat{Y}^{\mathcal{F}},
\]
where $\mathcal{W}$ denotes the fixed set of superquadrics and $\Pi_t$ its partition at step $t$ (Sec.~\ref{sec:workspace}). These components respectively abstract dense geometry into composable structural units, organize them into part-level identities, and localize the inferred identities on the original surface.

\begin{figure}[t]
    \centering
    \includegraphics[width=\linewidth]{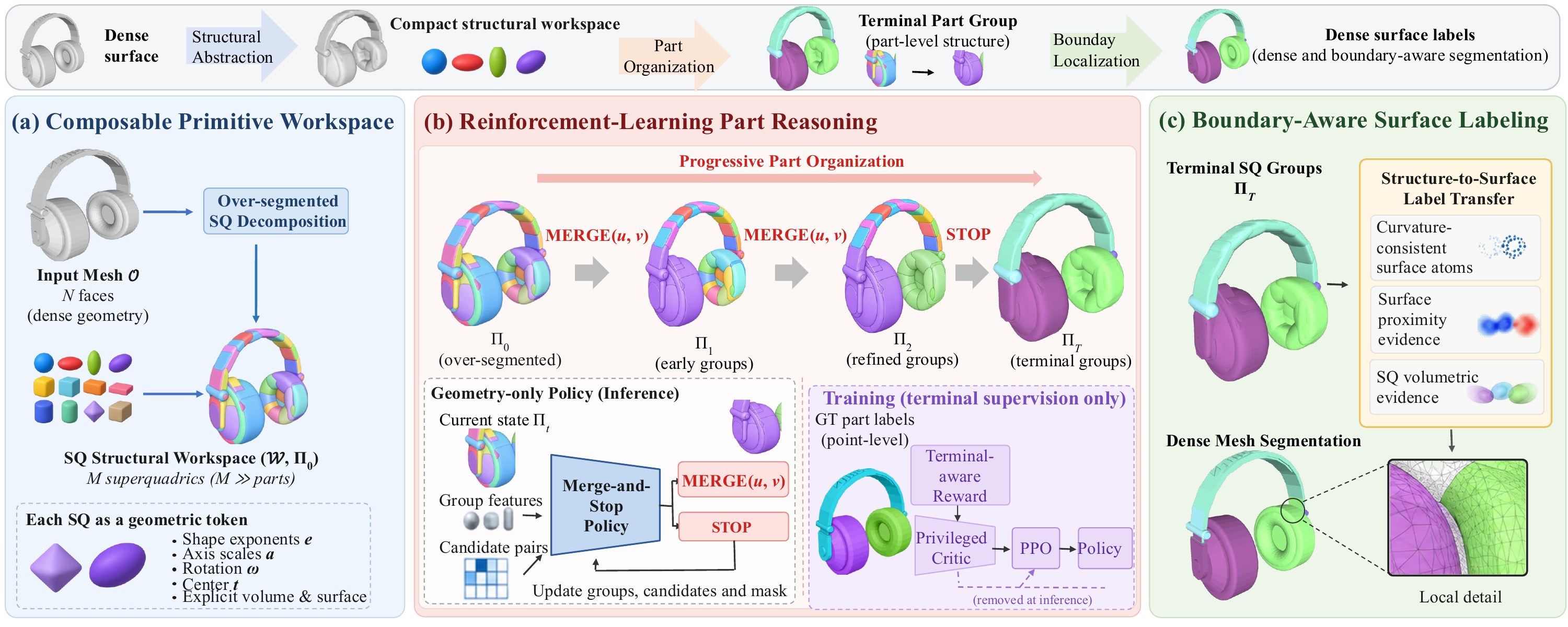}
    \caption{\textbf{Overview of RePart.}
    RePart addresses primitive-based 3D segmentation through three coupled stages that bridge dense geometry, part structure, and surface boundaries. (a) \textbf{Composable Primitive Workspace} abstracts the dense surface into compact, composable geometric units for structural reasoning. (b) \textbf{Reinforcement-Learning Part Reasoning} organizes these primitives into coherent part-level structure. (c) \textbf{Boundary-Aware Surface Labeling} localizes the inferred part identities back onto the original surface to recover dense, boundary-aware segmentation.}
    \label{fig:RePart-pipeline}
\end{figure}

\subsection{Composable Primitive Workspace}
\label{sec:workspace}

Structural abstraction requires compact units for group-level reasoning while preserving candidate part boundaries for subsequent organization. We therefore construct a fixed, intentionally over-segmented SQ basis whose primitives form a \emph{Composable Primitive Workspace}. SQs provide compact shape, pose, and volumetric support; over-segmentation allows one part to be composed from multiple primitives and preserves candidate boundaries for merge-only organization. Fixing SQ geometry and point correspondence makes alternative part organizations differ only in how the same primitives are grouped, turning structural inference into partition reasoning over a stable geometric basis. To construct this workspace, we normalize the mesh, compute its signed distance field, and initialize an intentionally over-segmented set with Marching-Primitives~\citep{liu2023marching}:
\begin{equation}
\mathcal W=\{p_1,\ldots,p_M\},
\end{equation} where \(p_j\) denotes the \(j\)-th fitted SQ primitive.

We next fix the correspondence between surface points and SQ primitives so that different part-organization hypotheses are evaluated over the same geometric basis. Let $d_{ij}=|d_{\mathrm{SQ}}(\vx_i;p_j)|$, using the standard radial signed-distance approximation defined in Appendix~\ref{app:sq-distance}. We compute soft support and fix each point's primitive assignment as
\begin{equation}
\alpha_{ij}=\frac{\exp(-d_{ij}/\beta)}{\sum_{\ell=1}^{M}\exp(-d_{i\ell}/\beta)},\qquad
z_i=\arg\max_j\alpha_{ij},\qquad
\beta=\operatorname{median}_{i,\ell}d_{i\ell}.
\end{equation}
Both SQ geometry and the point-to-SQ assignments $z_i$ remain fixed throughout reasoning. Starting from $\vg_0=(1,\ldots,M)$, the grouping vector $\vg_t=(g_{t,1},\ldots,g_{t,M})$ induces
\begin{equation}\label{eq:partition}
\Pi_t=\bigl\{\{\,j: g_{t,j}=u\,\}: u\in\mathcal{U}_t\bigr\},
\qquad
\mathcal{U}_t=\{g_{t,j}\}_{j=1}^{M},
\end{equation}
where $\mathcal U_t$ denotes the active group labels and $\Pi_t$ the induced SQ partition. $\mathcal W$ provides fixed composable units, while $\Pi_t$ captures their evolving part organization; with fixed $z_i$, all partitions are evaluated under the same point-to-SQ correspondence via $\hat y_{t,i}=g_{t,z_i}$.
\subsection{Reinforcement-Learning Part Reasoning}

\textbf{Merge-and-Stop Markov Decision Process.} The composable primitive workspace provides compact geometric units, but it does not determine their part-level organization. Because a semantic part may span multiple SQs, the model must determine which current groups should share a part identity and when further consolidation should stop. These decisions are coupled: each irreversible merge changes group geometry and the candidates available later, while supervision specifies the desired terminal partition rather than a unique merge sequence. We therefore formulate part reasoning as a merge-and-stop Markov decision process and optimize its trajectories with PPO, allowing decisions to be evaluated through their downstream effects on the resulting organization.

We model part reasoning as a finite-horizon Markov decision process $\mathcal M=(\mathcal S,\mathcal A,\mathcal P,\mathcal R)$, whose actions progressively reorganize the current partition until the policy selects \texttt{stop} or reaches the maximum horizon. 

\noindent\textbf{State ($\mathcal{S}$).} An episode starts from $(\mathcal W,\Pi_0)$, with underlying state $\tilde{s}_t=\{X,\mathcal{W},\vg_t\}$, where $\vg_t$ induces the current partition $\Pi_t$ as defined in Eq.~\eqref{eq:partition}.  The actor observes
\begin{equation}
o_t=\{X_s,\mathcal{W}, \Phi_t^{\mathrm{prim}},\Phi_t^{\mathrm{global}},
\Phi_t^{\mathrm{pair}},\veta_t\},
\end{equation}
where $X_s$ is sampled without replacement from $X$ at each encoding and $\veta_t$ is a binary action mask, which retains \texttt{stop} and valid merge candidates while suppressing unavailable actions. Rather than numeric group IDs, the actor uses geometry-derived descriptors: $\Phi_t^{\mathrm{prim}}$ describes each primitive and its current group, $\Phi_t^{\mathrm{global}}$ summarizes grouping progress, and $\Phi_t^{\mathrm{pair}}$ encodes candidate compatibility. The SQ parameters and point assignments remain fixed, while grouping-dependent descriptors and the candidate mask are recomputed after each merge.

\noindent\textbf{Action ($\mathcal{A}$).}
We form a dynamic candidate set $\mathcal{C}_t$ by connecting every active group to its $\kappa$ nearest groups using axis-aligned bounding-box separation, with zero distance for overlapping boxes, and then symmetrizing the relations. The available actions are

\begin{equation}
\begin{split}
\mathcal{A}(\tilde{s}_t)
=\{\texttt{stop}\}\cup
\{\texttt{merge}(u,v):(u,v)\in\mathcal{C}_t\}.
\end{split}
\end{equation}

\noindent\textbf{Transition ($\mathcal{P}$).}
For $a_t=\texttt{merge}(u,v)$, the transition deterministically relabels $v$ as $u$:
\begin{equation}
g_{t+1,j}=
\begin{cases}
u, & g_{t,j}=v,\\
g_{t,j}, & \text{otherwise}.
\end{cases}
\end{equation}
Thus, all SQs in group $v$ are reassigned to $u$, while all other assignments remain unchanged. Group-dependent descriptors, candidates, and the action mask are then recomputed. Resampling $X_s$ changes only the actor observation, not the underlying state.  Each merge reduces the group count by one. For $a_t=\texttt{stop}$, we keep $\vg_{t+1}=\vg_t$ and terminate the episode; the episode also terminates at the maximum horizon.

\noindent\textbf{Reward ($\mathcal{R}$).}
Training annotations specify the desired part organization but not a unique sequence of merges. We therefore supervise partition progress rather than individual actions using permutation-invariant overlap and group-count error. For ground-truth part $\mathcal Y_k=\{i:y_i=k\}$ and predicted group $\widehat{\mathcal{Y}}_{t,u}=\{i:\hat{y}_{t,i}=u\}$, we define 
\begin{equation}
m_t=\frac{1}{K}\sum_{k=1}^{K}
\max_{u\in\mathcal{U}_t}
\operatorname{IoU}(\mathcal{Y}_k,\widehat{\mathcal{Y}}_{t,u}),
E_t=\frac{|G_t-K|}{\max(K,1)},
\end{equation}
where $K$ is the number of ground-truth parts and $G_t=|\Pi_t|$. The overlap term follows permutation-invariant class-agnostic part
evaluation ~\citep{yang2024sampart3d,liu2025partfield}. 

The transition reward $\mathcal{R}(\tilde{s}_t,a_t,\tilde{s}_{t+1})=r_t$ is
\begin{equation}
r_t=\underbrace{m_{t+1}-m_t}_{\text{overlap gain}}+\lambda_g\underbrace{(E_t-E_{t+1})}_{\text{count correction}}+\mathbb{I}[a_t=\texttt{stop}]\underbrace{(b_{\mathrm{stop}}-\lambda_sE_{t+1})}_{\text{stop quality}}.
\end{equation}

The three terms respectively reward a better part organization, movement toward the correct group count, and a well-timed stop. Here, $\lambda_g,\lambda_s\geq0$ are weighting coefficients and $b_{\mathrm{stop}}$ is the base reward for an explicit stop. Labels $Y$ and $K$ provide training rewards but are unavailable to the deployed actor.

\textbf{Geometry-Only Grouping Policy.}
Although part annotations supervise training, the deployed actor must infer part organization from geometry alone. We therefore encode evidence at three complementary levels: surface points provide object-level context, SQ descriptors characterize the current groups, and pair descriptors describe the geometry of candidate merges. 
Centered, radius-normalized points are encoded by a shared MLP with max--mean pooling; separate MLPs encode SQ and primitive descriptors, which are aggregated within each group by masked mean pooling.

As groups evolve, their descriptors aggregate the geometry of all constituent SQs, while a fixed catalog representative maintains consistent action indexing. The merge and stop heads combine the shared geometric context with action-specific descriptors and a geometry-predicted estimate $\hat\nu_t$ of the target group-count ratio. The latter is supervised during training but requires no labels at inference.
Given catalog logits $h_\theta(o_t)$, the masked policy is
\begin{equation}
\pi_\theta(a_t\mid o_t)=
\operatorname{softmax}\!\left(h_\theta(o_t)+\log\veta_t\right).
\end{equation}
The logarithmic mask assigns $-\infty$ to invalid actions, so the softmax normalizes over $\mathcal A(\tilde s_t)$ only.

\textbf{PPO Optimization.}
We optimize the merge-and-stop policy with PPO~\citep{schulman2017proximal}
using the trajectory rewards above, so each action is credited according to
its downstream effect on subsequent partitions and the final stopping decision:
\begin{equation}
\mathcal{L}_{\mathrm{PPO}}=-\mathbb{E}_t\left[
\min\left(\rho_t(\theta)\hat{A}_t,
\operatorname{clip}(\rho_t(\theta),1-\epsilon,1+\epsilon)\hat{A}_t\right)
\right],
\end{equation}
where $\rho_t$ is the policy ratio, $\hat A_t$ the estimated advantage, and $\epsilon$ the clipping radius.

The critic uses stronger training-time supervision than the actor, combining a detached actor representation with label-derived partition statistics to estimate state value and terminal quality. Detachment prevents these privileged signals from entering the actor representation. The full training objective further includes value, entropy, terminal-prediction, and granularity-prediction losses. The resulting terminal partition $\Pi_T$ encodes the inferred part-level organization and is subsequently passed to Boundary-Aware Surface Labeling in Sec.~\ref{sec:recovery}.

\subsection{Boundary-Aware Surface Labeling}
\label{sec:recovery}
The terminal partition $\Pi_T$ determines which SQs share the same part identity, but their primitive boundaries only approximate the original surface and therefore do not directly localize precise mesh boundaries. We perform boundary localization on the mesh itself through \emph{Boundary-Aware Surface Labeling}. Curvature-consistent atoms constrain where labels may change, while grouped SQ surface and volumetric evidence determine which part identity each atom receives.

Each terminal group combines the geometric support of its constituent SQs, allowing one part to span multiple primitives. For $u\in\mathcal U_T$, we define the grouped surface samples as $\mathcal B_u=\bigcup_{j:g_{T,j}=u}\mathcal B(p_j)$, where $\mathcal B(p_j)$ denotes samples from the analytic surface of SQ $p_j$.
To localize boundaries without fragmenting coherent surface regions through independent face-wise assignments,  we cut high-dihedral adjacencies to form curvature-consistent atoms $\{\Omega_q\}_{q=1}^{Q}$, each of which receives a single group label. For face centroid $\vc_f$, we compute
\begin{equation}
\delta_{fu}=\min_{\vb\in\mathcal B_u}\|\vc_f-\vb\|_2,\qquad
\tilde u_f=\arg\min_u\delta_{fu},\qquad
\delta_f=\delta_{f\tilde u_f}.
\end{equation}
Each atom then aggregates valid proximity votes:
\begin{equation}
V_{qu}=\sum_{f\in\Omega_q}\mathbb I[\tilde u_f=u]\mathbb I[\delta_f\leq\tau_d],\qquad
\hat u_q=\arg\max_u V_{qu}.
\end{equation}
We accept this assignment when at least one valid vote exists and at most a fraction $\tau_o$ of the atom's faces exceed $\tau_d$.  Surface proximity alone can leave regions unresolved that lie within a group's volumetric support but remain far from its fitted surface. For rejected atoms, we therefore evaluate signed radial support using Eq.~\eqref{eq:dsq}:
\begin{equation}
D_u(\vc_f)=\min_{j:g_{T,j}=u}d_{\mathrm{SQ}}(\vc_f;p_j).
\end{equation}
A rejected atom is assigned to group $u$ when at least half of its face centroids satisfy $D_u(\vc_f)\leq\tau_c$; otherwise, we use the group receiving the most unconstrained nearest-surface votes. Each face then inherits its atom label, $\hat y_f^{\mathcal F}=\hat u_q$ for $f\in\Omega_q$.
This boundary-aware labeling preserves the inferred part-level organization while allowing final label transitions to follow the original mesh geometry rather than primitive interfaces.
\section{Experiments}

\subsection{Experimental Setup}

\textbf{Datasets and splits.}
We train RePart on PartNet~\citep{mo2019partnet}, grouping relevant categories into furniture, containers, and daily-use objects and sampling 300 shapes per group. We test on 200 total corresponding-category shapes from PartNet and 200 total from 3DCoMPaT++~\citep{slim20253dcompat++}. The latter is used only for testing, without target-dataset fine-tuning.

\textbf{Training.}
We first optimize one category-specialized policy per group over 200 epochs. The resulting parameters initialize a joint fine-tuning stage on the union of the three training subsets for 100 epochs. All experiments are run on a single NVIDIA RTX 3090 GPU. Unless otherwise specified, RePart denotes this jointly fine-tuned model. Appendix Figure~\ref{fig:training-curves} reports the training rewards; to prevent reward-scale differences across category groups from joint training, we normalize advantages separately within each category, average its loss, and then macro-average the category losses.

\textbf{Baselines and metrics.}
We compare with Marching-Primitives (MPS)~\citep{liu2023marching}, PartField~\citep{liu2025partfield}, P3-SAM~\citep{ma2025p3}, SAMPart3D~\citep{yang2024sampart3d}, and PrimitiveAnything (PriMAny)~\citep{ye2025primitiveanything}. We report mean intersection-over-union (mIoU)~\citep{yang2024sampart3d,liu2025partfield} to measure overlap between matched predicted and ground-truth parts; Rand Index (RI)~\citep{rand1971objective} to measure whether pairs of faces are consistently assigned to the same or different parts; Variation of Information (VoI)~\citep{meilua2007comparing} to quantify the information difference between two partitions, including splitting and merging errors; and Segmentation Covering (SC)~\citep{arbelaez2010contour} to measure how well predicted segments cover ground-truth regions. Following the PartField protocol, we report mIoU across its candidate granularities, with ground-truth labels used only for evaluation-time granularity selection.

\begin{table}[t]
\caption{Category-group mIoU on PartNet and 3DCoMPaT++. RePart uses joint fine-tuning; \textbf{bold} and \underline{underlined} values denote the best and second-best results.}
\label{tab:group-miou}
\centering
\small
\setlength{\tabcolsep}{7pt}
\begin{tabular}{lcccccc}
\toprule
& \multicolumn{3}{c}{PartNet} & \multicolumn{3}{c}{3DCoMPaT++} \\
\cmidrule(lr){2-4}\cmidrule(lr){5-7}
Method & Furn. & Cont. & Daily & Furn. & Cont. & Daily \\
\midrule
MPS       & 0.1591 & 0.5313 & 0.2169 & 0.3193 & 0.1144 & 0.2008 \\
PartField & \underline{0.4098} & \underline{0.6418} & 0.4969 & 0.2438 & 0.4127 & 0.4018 \\
P3-SAM    & 0.3243 & 0.5663 & \underline{0.5411} & \best{0.4147} & \underline{0.5110} & \best{0.6003} \\
SAMPart3D & 0.2520 & 0.5111 & 0.4899 & 0.3313 & 0.4544 & \underline{0.4720} \\
PriMAny   & 0.2255 & 0.2067 & 0.3032 & 0.2419 & 0.1964 & 0.3236 \\
\midrule
\textbf{RePart (Ours)} & \best{0.4247} & \best{0.6456} & \best{0.5432} & \underline{0.3703} & \best{0.5149} & 0.4607 \\
\bottomrule
\end{tabular}
\end{table}

\subsection{Main Results}

\textbf{Quantitative Results.}
 Tables~\ref{tab:group-miou} and~\ref{tab:partition-metrics} show consistently strong \emph{in-domain performance} for RePart on PartNet across both category-level mIoU and partition-level metrics. RePart achieves the highest mIoU across all three category groups, reaching 0.4247 on furniture, 0.6456 on containers, and 0.5432 on daily-use objects. It also obtains the strongest aggregate results across all four partition metrics. Compared with the strongest competing results, RePart improves mIoU from 0.5249 to 0.5460 and SC from 0.6502 to 0.6891, while reducing VoI from 1.0213 to 0.8471. These results establish that the complete pipeline produces accurate and structurally coherent partitions across complementary evaluation criteria.

On 3DCoMPaT++, RePart demonstrates cross-dataset transfer without fine-tuning, achieving the highest container mIoU (0.5149), while P3-SAM performs better on furniture and daily-use objects. Across all test shapes, RePart ranks first in RI (0.7575) and SC (0.5790), and second in both mIoU and VoI. Notably, RePart does not lead in transferred mIoU, but retains the strongest RI and SC. The transferred model therefore retains strong pairwise and region-level partition consistency despite the dataset shift. On six PartNet categories excluded from training, RePart further achieves the best VoI and SC and the second-best average mIoU; full results are provided in Appendix~\ref{sec:appendix-unseen}.

\begin{figure}[!t]
    \centering
    \includegraphics[width=\linewidth,height=0.82\textheight,keepaspectratio]{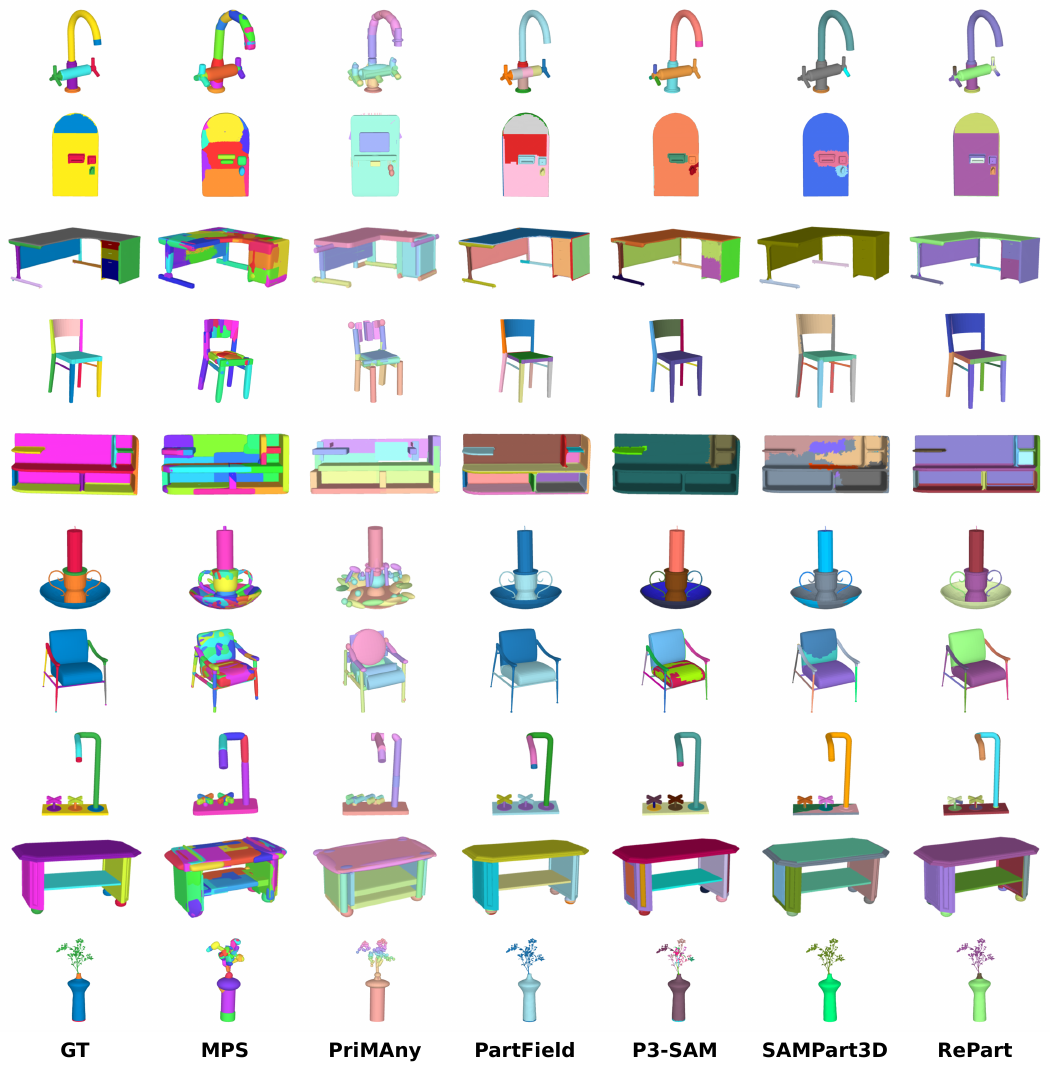}
    \caption{Qualitative comparison on five PartNet shapes (top five rows) and five 3DCoMPaT++ shapes (bottom five rows). RePart reduces the fragmented MPS decomposition and produces coherent part partitions across both datasets.}
    \label{fig:qualitative-results}
\end{figure}

\textbf{Qualitative Results.}
Figure~\ref{fig:qualitative-results} compares segmentation results on five PartNet shapes and five 3DCoMPaT++ shapes. On PartNet, RePart produces coherent part partitions while reducing the strong fragmentation visible in the initial MPS decomposition. Compared with the other baselines, it also preserves localized structures such as thin supports, attachments, and articulated components in the shown examples. On 3DCoMPaT++, RePart retains similarly coherent part organization without target-dataset fine-tuning, consistent with the cross-dataset results in Table~\ref{tab:partition-metrics}.  

\begin{table*}[t]
\centering

\begin{minipage}{0.98\textwidth}
\centering
\caption{Average partition quality on PartNet / 3DCoMPaT++.}
\label{tab:partition-metrics}
\small
\setlength{\tabcolsep}{5.3pt}

\begin{tabular}{lcccccccc}
\toprule
& \multicolumn{4}{c}{PartNet} & \multicolumn{4}{c}{3DCoMPaT++} \\
\cmidrule(lr){2-5}\cmidrule(lr){6-9}
Method
& mIoU$\uparrow$ & RI$\uparrow$ & VoI$\downarrow$ & SC$\uparrow$
& mIoU$\uparrow$ & RI$\uparrow$ & VoI$\downarrow$ & SC$\uparrow$ \\
\midrule
MPS
& 0.3169 & 0.5266 & \underline{1.0213} & 0.5267
& 0.2125 & 0.4371 & \best{1.2043} & 0.4371 \\

PartField
& \underline{0.5249} & \underline{0.7877} & 1.0810 & \underline{0.6502}
& 0.3483 & 0.6914 & 1.2998 & 0.5302 \\

P3-SAM
& 0.4858 & 0.6966 & 1.2238 & 0.5833
& \best{0.5003} & \underline{0.7455} & 1.3783 & \underline{0.5606} \\

SAMPart3D
& 0.4229 & 0.6999 & 1.1996 & 0.5862
& 0.4144 & 0.7089 & 1.5508 & 0.5304 \\

PriMAny
& 0.2440 & 0.5231 & 2.6506 & 0.2476
& 0.2477 & 0.5932 & 2.4509 & 0.2796 \\
\midrule

\textbf{RePart (Ours)}
& \best{0.5460} & \best{0.7965} & \best{0.8471} & \best{0.6891}
& \underline{0.4476} & \best{0.7575} & \underline{1.2838} & \best{0.5790} \\
\bottomrule
\end{tabular}
\end{minipage}

\vspace{15pt}

\begin{minipage}{0.98\textwidth}
\centering
\caption{Category-level mIoU of category-specific and jointly fine-tuned policies.}
\label{tab:ablation-specialized-joint}
\small
\setlength{\tabcolsep}{5.3pt}

\begin{tabular}{lcccccccc}
\toprule
& \multicolumn{4}{c}{PartNet} & \multicolumn{4}{c}{3DCoMPaT++} \\
\cmidrule(lr){2-5}\cmidrule(lr){6-9}
Policy
& Furn. & Cont. & Daily & Avg.
& Furn. & Cont. & Daily & Avg. \\
\midrule

Category-specific
& \best{0.4280} & \best{0.6515} & 0.5075 & 0.5375
& 0.3699 & 0.4517 & \best{0.4646} & 0.4255 \\

Staged joint
& 0.4247 & 0.6456 & \best{0.5432} & \best{0.5460}
& \best{0.3703} & \best{0.5149} & 0.4607 & \best{0.4476} \\
\bottomrule
\end{tabular}
\end{minipage}

\vspace{15pt}

\begin{minipage}{0.98\textwidth}
\centering
\caption{Optimization-objective ablations on furniture models. Learned variants use the same SQ state/action representation and actor parameterization.}
\label{tab:ablation-training-policy}
\small
\setlength{\tabcolsep}{4.2pt}

\begin{tabular}{lcccccccc}
\toprule
& \multicolumn{4}{c}{PartNet furniture}
& \multicolumn{4}{c}{3DCoMPaT++ furniture} \\
\cmidrule(lr){2-5}\cmidrule(lr){6-9}
Policy
& mIoU$\uparrow$ & RI$\uparrow$ & VoI$\downarrow$ & SC$\uparrow$
& mIoU$\uparrow$ & RI$\uparrow$ & VoI$\downarrow$ & SC$\uparrow$ \\
\midrule

Greedy-BC
& 0.3757 & 0.7301 & 1.4123 & 0.5284
& 0.3442 & 0.7494 & 1.9009 & 0.4226 \\

Contextual bandit
& 0.2591 & 0.4613 & 1.5346 & 0.4614
& 0.2144 & 0.3502 & 1.7428 & 0.3002 \\

Pairwise classifier
& 0.3275 & 0.6362 & 1.3656 & 0.4971
& 0.2465 & 0.5499 & 1.7450 & 0.3700 \\

Myopic-PPO
& 0.2579 & 0.4982 & 1.3813 & 0.4420
& 0.1678 & 0.3354 & \best{1.6703} & 0.2980 \\

RePart-S
& \best{0.4280} & \best{0.7804} & \best{1.2997} & \best{0.5751}
& \best{0.3699} & \best{0.7846} & 1.8601 & \best{0.4452} \\
\bottomrule
\end{tabular}
\end{minipage}

\end{table*}

\subsection{Ablation Studies}

\textbf{Category specialization and joint fine-tuning.}
We first compare the three category-specialized policies with the jointly fine-tuned policy. Table~\ref{tab:ablation-specialized-joint} reports category-level mIoU. Joint fine-tuning improves the average from 0.5375 to 0.5460 on PartNet and from 0.4255 to 0.4476 on 3DCoMPaT++. Its gains on PartNet daily-use objects and 3DCoMPaT++ furniture and containers indicate that sharing structural experience across category groups improves transfer.

\textbf{Reinforcement-Learning Part Reasoning.}
We compare RePart-S, trained on 300 PartNet furniture shapes, with agglomerative policies sharing the same actor architecture, geometry-only SQ state, merge-and-stop action space, action mask, training data, and environment. Following Learning to Group~\citep{luo2020learning}, we adapt a contextual bandit and pairwise merge classifier. The former optimizes immediate transition rewards, while the latter learns from counterfactual one-step rewards over legal actions. Myopic-PPO sets $\gamma=0$ and $\lambda_{\mathrm{GAE}}=0$, removing temporal bootstrapping and future rewards, while Greedy-BC uses locally optimal merge trajectories as supervision. These controls isolate the effect of optimizing multi-step returns.

Table~\ref{tab:ablation-training-policy} shows that the contextual bandit and Myopic-PPO struggle to learn grouping under this environment, indicating that immediate rewards alone provide insufficient guidance for the irreversible merge sequence. The pairwise classifier and Greedy-BC make stronger local decisions, but both yield lower mIoU, RI, and SC than full PPO. 
These results support optimizing multi-step returns for SQ grouping rather than treating merges as independent or locally optimal decisions.

\textbf{Boundary-Aware Surface Labeling.}
We isolate Boundary-Aware Surface Labeling by fixing the terminal SQ groups and varying only their transfer to mesh faces. Appendix Table~\ref{tab:appendix-projection} shows that curvature-consistent atoms substantially improve mIoU over direct facewise nearest-SQ assignment across all three PartNet groups, while SQ volumetric evidence further resolves regions left uncertain by surface proximity and reduces fallback usage. These results support separating \emph{Semantic Organization} from \emph{Boundary Localization}.

\section{Conclusion and Discussion}

We introduced \textbf{RePart}, a framework for \emph{Semantic Organization} in 3D part segmentation that treats geometric primitives as reasoning units rather than final parts. RePart formulates primitive grouping as a finite-horizon merge-and-stop process, using trajectory-level reinforcement learning to optimize decisions by their downstream effects on the final partition, while Boundary-Aware Surface Labeling maps inferred part identities back to the mesh. Experiments on PartNet and 3DCoMPaT++ support our core claim that multi-step reinforcement reasoning effectively organizes a fixed primitive workspace into coherent part-level structure, rather than primitive decomposition alone yielding semantic parts. Comparisons with local and myopic objectives further show the benefit of optimizing sequential decisions. Compact geometric primitives thus provide an intermediate reasoning space between dense geometry and semantic structure. RePart separates structural abstraction, part organization, and boundary localization.
The fixed MPS-derived workspace provides a stable basis for sequential reasoning, but also ties performance to the quality of the initial decomposition. Our current study focuses on 900 shapes across three category groups; extending the action space and training coverage may further improve flexibility and generalization.

\flushbottom

\bibliography{RePart_refs}

@article{barr1981superquadrics,
  title={Superquadrics and angle-preserving transformations},
  author={Barr, Alan H},
  journal={IEEE Computer graphics and Applications},
  volume={1},
  number={01},
  pages={11--23},
  year={1981},
  publisher={IEEE Computer Society}
}

@inproceedings{fedele2025superdec,
  title={Superdec: 3d scene decomposition with superquadric primitives},
  author={Fedele, Elisabetta and Sun, Boyang and Guibas, Leonidas and Pollefeys, Marc and Engelmann, Francis},
  booktitle={2025 IEEE/CVF International Conference on Computer Vision (ICCV)},
  pages={24625--24635},
  year={2025},
  organization={IEEE}
}

@inproceedings{liu2025partfield,
  title={Partfield: Learning 3d feature fields for part segmentation and beyond},
  author={Liu, Minghua and Uy, Mikaela Angelina and Xiang, Donglai and Su, Hao and Fidler, Sanja and Sharp, Nicholas and Gao, Jun},
  booktitle={2025 IEEE/CVF International Conference on Computer Vision (ICCV)},
  pages={9704--9715},
  year={2025},
  organization={IEEE}
}

@article{li20233d,
  title={3D reconstruction based on hierarchical reinforcement learning with transferability},
  author={Li, Lan and He, Fazhi and Fan, Rubin and Fan, Bo and Yan, Xiaohu},
  journal={Integrated Computer-Aided Engineering},
  volume={30},
  number={4},
  pages={327--339},
  year={2023},
  publisher={SAGE Publications Sage UK: London, England}
}

@inproceedings{liu2023marching,
  title={Marching-primitives: Shape abstraction from signed distance function},
  author={Liu, Weixiao and Wu, Yuwei and Ruan, Sipu and Chirikjian, Gregory S},
  booktitle={2023 IEEE/CVF Conference on Computer Vision and Pattern Recognition (CVPR)},
  pages={8771--8780},
  year={2023},
  organization={IEEE}
}

@inproceedings{mo2019partnet,
  title={Partnet: A large-scale benchmark for fine-grained and hierarchical part-level 3d object understanding},
  author={Mo, Kaichun and Zhu, Shilin and Chang, Angel X and Yi, Li and Tripathi, Subarna and Guibas, Leonidas J and Su, Hao},
  booktitle={2019 IEEE/CVF Conference on Computer Vision and Pattern Recognition (CVPR)},
  pages={909--918},
  year={2019},
  organization={IEEE}
}

@inproceedings{paschalidou2021neural,
  title={Neural parts: Learning expressive 3d shape abstractions with invertible neural networks},
  author={Paschalidou, Despoina and Katharopoulos, Angelos and Geiger, Andreas and Fidler, Sanja},
  booktitle={2021 IEEE/CVF Conference on Computer Vision and Pattern Recognition (CVPR)},
  pages={3203--3214},
  year={2021},
  organization={IEEE}
}

@article{schulman2017proximal,
  title={Proximal policy optimization algorithms},
  author={Schulman, John and Wolski, Filip and Dhariwal, Prafulla and Radford, Alec and Klimov, Oleg},
  journal={arXiv preprint arXiv:1707.06347},
  year={2017}
}

@article{solina1990recovery,
  title={Recovery of parametric models from range images: The case for superquadrics with global deformations},
  author={Solina, Franc and Bajcsy, Ruzena},
  journal={IEEE transactions on pattern analysis and machine intelligence},
  volume={12},
  number={2},
  pages={131--147},
  year={1990},
  publisher={IEEE}
}

@inproceedings{wang2025light,
  title={Light-sq: Structure-aware shape abstraction with superquadrics for generated meshes},
  author={Wang, Yuhan and Chen, Weikai and Hu, Zeyu and Zhang, Runze and Yin, Yingda and Wu, Ruoyu and Luo, Keyang and Qian, Shengju and Ma, Yiyan and Li, Hongyi and others},
  booktitle={Proceedings of the SIGGRAPH Asia 2025 Conference Papers},
  pages={1--11},
  year={2025}
}

@article{slim20253dcompat++,
  title={3dcompat++: An improved large-scale 3d vision dataset for compositional recognition},
  author={Slim, Habib and Li, Xiang and Li, Yuchen and Ahmed, Mahmoud and Ayman, Mohamed and Upadhyay, Ujjwal and Abdelreheem, Ahmed and Prajapati, Arpit and Pothigara, Suhail and Wonka, Peter and others},
  journal={IEEE Transactions on Pattern Analysis and Machine Intelligence},
  year={2025},
  publisher={IEEE}
}

@article{ma2025p3,
  title={P3-sam: Native 3d part segmentation},
  author={Ma, Changfeng and Li, Yang and Yan, Xinhao and Xu, Jiachen and Yang, Yunhan and Wang, Chunshi and Zhao, Zibo and Guo, Yanwen and Chen, Zhuo and Guo, Chunchao},
  journal={arXiv preprint arXiv:2509.06784},
  year={2025}
}

@article{yang2024sampart3d,
  title={Sampart3d: Segment any part in 3d objects},
  author={Yang, Yunhan and Huang, Yukun and Guo, Yuan-Chen and Lu, Liangjun and Wu, Xiaoyang and Lam, Edmund Y and Cao, Yan-Pei and Liu, Xihui},
  journal={arXiv preprint arXiv:2411.07184},
  year={2024}
}

@inproceedings{ye2025primitiveanything,
  title={Primitiveanything: Human-crafted 3d primitive assembly generation with auto-regressive transformer},
  author={Ye, Jingwen and He, Yuze and Zhou, Yanning and Zhu, Yiqin and Xiao, Kaiwen and Liu, Yong-Jin and Yang, Wei and Han, Xiao},
  booktitle={Proceedings of the Special Interest Group on Computer Graphics and Interactive Techniques Conference Conference Papers},
  pages={1--12},
  year={2025}
}

@article{rand1971objective,
  title={Objective criteria for the evaluation of clustering methods},
  author={Rand, William M},
  journal={Journal of the American Statistical association},
  volume={66},
  number={336},
  pages={846--850},
  year={1971},
  publisher={Taylor \& Francis}
}

@article{meilua2007comparing,
  title={Comparing clusterings—an information based distance},
  author={Meil{\u{a}}, Marina},
  journal={Journal of multivariate analysis},
  volume={98},
  number={5},
  pages={873--895},
  year={2007},
  publisher={Elsevier}
}

@article{arbelaez2010contour,
  title={Contour detection and hierarchical image segmentation},
  author={Arbelaez, Pablo and Maire, Michael and Fowlkes, Charless and Malik, Jitendra},
  journal={IEEE transactions on pattern analysis and machine intelligence},
  volume={33},
  number={5},
  pages={898--916},
  year={2010},
  publisher={IEEE}
}

@inproceedings{liu2023partslip,
  title={Partslip: Low-shot part segmentation for 3d point clouds via pretrained image-language models},
  author={Liu, Minghua and Zhu, Yinhao and Cai, Hong and Han, Shizhong and Ling, Zhan and Porikli, Fatih and Su, Hao},
  booktitle={2023 IEEE/CVF Conference on Computer Vision and Pattern Recognition (CVPR)},
  pages={21736--21746},
  year={2023},
  organization={IEEE}
}

@inproceedings{umam2024partdistill,
  title={Partdistill: 3d shape part segmentation by vision-language model distillation},
  author={Umam, Ardian and Yang, Cheng-Kun and Chen, Min-Hung and Chuang, Jen-Hui and Lin, Yen-Yu},
  booktitle={2024 IEEE/CVF Conference on Computer Vision and Pattern Recognition (CVPR)},
  pages={3470--3479},
  year={2024},
  organization={IEEE}
}

@article{tang2024segment,
  title={Segment any mesh},
  author={Tang, George and Zhao, William and Ford, Logan and Benhaim, David and Zhang, Paul},
  journal={arXiv preprint arXiv:2408.13679},
  year={2024}
}

@inproceedings{ma2025find,
  title={Find any part in 3d},
  author={Ma, Ziqi and Yue, Yisong and Gkioxari, Georgia},
  booktitle={2025 IEEE/CVF International Conference on Computer Vision (ICCV)},
  pages={1--10},
  year={2025},
  organization={IEEE}
}

@inproceedings{zhu2026partsam,
  title={Partsam: A scalable promptable part segmentation model trained on native 3d data},
  author={Zhu, Zhe and Wan, Le and Xu, Rui and Zhang, Yiheng and Chen, Honghua and Dou, Zhiyang and Lin, Cheng and Liu, Yuan and Wei, Mingqiang},
  booktitle={International Conference on Learning Representations},
  volume={2026},
  pages={132469--132493},
  year={2026}
}

@article{hadgi2026patchalign3d,
  title={Patchalign3d: Local feature alignment for dense 3d shape understanding},
  author={Hadgi, Souhail and Gong, Bingchen and Sundararaman, Ramana and Pierson, Emery and Li, Lei and Wonka, Peter and Ovsjanikov, Maks},
  journal={arXiv preprint arXiv:2601.02457},
  year={2026}
}

@inproceedings{wu2022primitive,
  title={Primitive-based shape abstraction via nonparametric bayesian inference},
  author={Wu, Yuwei and Liu, Weixiao and Ruan, Sipu and Chirikjian, Gregory S},
  booktitle={European Conference on Computer Vision},
  pages={479--495},
  year={2022},
  organization={Springer}
}

@article{monnier2023differentiable,
  title={Differentiable blocks world: Qualitative 3d decomposition by rendering primitives},
  author={Monnier, Tom and Austin, Jake and Kanazawa, Angjoo and Efros, Alexei and Aubry, Mathieu},
  journal={Advances in Neural Information Processing Systems},
  volume={36},
  pages={5791--5807},
  year={2023}
}

@inproceedings{liu2023deformer,
  title={Deformer: Integrating transformers with deformable models for 3d shape abstraction from a single image},
  author={Liu, Di and Yu, Xiang and Ye, Meng and Zhangli, Qilong and Li, Zhuowei and Zhang, Zhixing and Metaxas, Dimitris N},
  booktitle={2023 IEEE/CVF International Conference on Computer Vision (ICCV)},
  pages={14190--14200},
  year={2023},
  organization={IEEE}
}

@article{niu2022rim,
  title={Rim-net: Recursive implicit fields for unsupervised learning of hierarchical shape structures},
  author={Niu, Chengjie and Li, Manyi and Xu, Kai and Zhang, Hao},
  journal={2022 IEEE/CVF Conference on Computer Vision and Pattern Recognition (CVPR)},
  year={2022},
  pages={11769-11778}
}

@inproceedings{du2024generative,
  title={Generative 3d part assembly via part-whole-hierarchy message passing},
  author={Du, Bi'an and Gao, Xiang and Hu, Wei and Liao, Renjie},
  booktitle={2024 IEEE/CVF Conference on Computer Vision and Pattern Recognition (CVPR)},
  pages={20850--20859},
  year={2024},
  organization={IEEE}
}

@inproceedings{ruan2025learning,
  title={Learning Partonomic 3D Reconstruction from Image Collections},
  author={Ruan, Xiaoqian and Yu, Pei and Jia, Dian and Park, Hyeonjeong and Xiong, Peixi and Tang, Wei},
  booktitle={2025 IEEE/CVF Conference on Computer Vision and Pattern Recognition (CVPR)},
  pages={26734--26744},
  year={2025},
  organization={IEEE}
}

@inproceedings{liu2025wir3d,
  title={Wir3d: Visually-informed and geometry-aware 3d shape abstraction},
  author={Liu, Richard and Fu, Daniel and Tan, Noah and Lang, Itai and Hanocka, Rana},
  booktitle={2025 IEEE/CVF International Conference on Computer Vision (ICCV)},
  pages={14810--14821},
  year={2025},
  organization={IEEE}
}

@inproceedings{su2026s,
  title={S$^{2}$AM3D: Scale-controllable Part Segmentation of 3D Point Clouds},
  author={Su, Han and Huang, Tianyu and Wan, Zichen and Wu, Xiaohe and Zuo, Wangmeng},
  booktitle={Proceedings of the IEEE/CVF Conference on Computer Vision and Pattern Recognition},
  pages={14357--14366},
  year={2026}
}

@inproceedings{deng2026geosam2,
  title={Geosam2: Unleashing the power of sam2 for 3d part segmentation},
  author={Deng, Ken and Yang, Yunhan and Sun, Jingxiang and Liu, Xihui and Liu, Yebin and Liang, Ding and Cao, Yan-Pei},
  booktitle={Proceedings of the IEEE/CVF Conference on Computer Vision and Pattern Recognition},
  pages={6367--6376},
  year={2026}
}

@article{von2026gaussiangpt,
  title={GaussianGPT: Towards Autoregressive 3D Gaussian Scene Generation},
  author={von L{\"u}tzow, Nicolas and R{\"o}ssle, Barbara and Schmid, Katharina and Nie{\ss}ner, Matthias},
  journal={arXiv preprint arXiv:2603.26661},
  year={2026}
}

@article{luo2020learning,
      title={Learning to Group: A Bottom-Up Framework for 3D Part Discovery in Unseen Categories},
      author={Tiange Luo and Kaichun Mo and Zhiao Huang and Jiarui Xu and Siyu Hu and Liwei Wang and Hao Su},
    booktitle={International Conference on Learning Representations (ICLR)},
    year={2020},
    url={https://iclr.cc/virtual_2020/poster_rkl8dlHYvB.html}
}

@article{leonardis1997superquadrics,
  author  = {Leonardis, Ale{\v{s}} and Jakli{\v{c}}, Ale{\v{s}} and Solina, Franc},
  title   = {Superquadrics for Segmenting and Modeling Range Data},
  journal = {IEEE Transactions on Pattern Analysis and Machine Intelligence},
  volume  = {19},
  number  = {11},
  pages   = {1289--1295},
  year    = {1997},
  doi     = {10.1109/34.632988}
}

@article{attene2006hierarchical,
  author  = {Attene, Marco and Falcidieno, Bianca and Spagnuolo, Michela},
  title   = {Hierarchical mesh segmentation based on fitting primitives},
  journal = {The Visual Computer},
  volume  = {22},
  number  = {3},
  pages   = {181--193},
  year    = {2006},
  doi     = {10.1007/s00371-006-0375-x}
}

@inproceedings{brehmer2020hierarchical,
  title     = {Hierarchical clustering in particle physics through reinforcement learning},
  author    = {Brehmer, Johann and Macaluso, Sebastian and Pappadopulo, Duccio and Cranmer, Kyle},
  booktitle = {Machine Learning and the Physical Sciences Workshop at NeurIPS},
  year      = {2020},
  eprint    = {2011.08191},
  archivePrefix = {arXiv},
  url       = {https://arxiv.org/abs/2011.08191}
}

@inproceedings{qi2024shapellm,
  title={Shapellm: Universal 3d object understanding for embodied interaction},
  author={Qi, Zekun and Dong, Runpei and Zhang, Shaochen and Geng, Haoran and Han, Chunrui and Ge, Zheng and Yi, Li and Ma, Kaisheng},
  booktitle={European Conference on Computer Vision},
  pages={214--238},
  year={2024},
  organization={Springer}
}

@article{ganeshan2024parsel,
  title={ParSEL: Parameterized shape editing with language},
  author={Ganeshan, Aditya and Huang, Ryan and Xu, Xianghao and Jones, R Kenny and Ritchie, Daniel},
  journal={ACM Transactions on Graphics (TOG)},
  volume={43},
  number={6},
  pages={1--14},
  year={2024},
  publisher={ACM New York, NY, USA}
}

@inproceedings{geng2023gapartnet,
  title={Gapartnet: Cross-category domain-generalizable object perception and manipulation via generalizable and actionable parts},
  author={Geng, Haoran and Xu, Helin and Zhao, Chengyang and Xu, Chao and Yi, Li and Huang, Siyuan and Wang, He},
  booktitle={2023 IEEE/CVF Conference on Computer Vision and Pattern Recognition (CVPR)},
  pages={7081--7091},
  year={2023},
  organization={IEEE}
}

@inproceedings{paschalidou2019superquadrics,
  title={Superquadrics revisited: Learning 3d shape parsing beyond cuboids},
  author={Paschalidou, Despoina and Ulusoy, Ali Osman and Geiger, Andreas},
  booktitle={2019 IEEE/CVF Conference on Computer Vision and Pattern Recognition (CVPR)},
  pages={10336--10345},
  year={2019},
  organization={IEEE}
}

@inproceedings{li2019supervised,
  title={Supervised fitting of geometric primitives to 3d point clouds},
  author={Li, Lingxiao and Sung, Minhyuk and Dubrovina, Anastasia and Yi, Li and Guibas, Leonidas J},
  booktitle={2019 IEEE/CVF conference on computer vision and pattern recognition (CVPR)},
  pages={2647--2655},
  year={2019},
  organization={IEEE}
}

@article{puterman1990markov,
  title={Markov decision processes},
  author={Puterman, Martin L},
  journal={Handbooks in operations research and management science},
  volume={2},
  pages={331--434},
  year={1990},
  publisher={Elsevier}
}

@book{sutton1998reinforcement,
  title={Reinforcement learning: An introduction},
  author={Sutton, Richard S and Barto, Andrew G and Barto, Andrew},
  volume={1},
  number={1},
  year={1998},
  publisher={MIT press Cambridge}
}
\bibliographystyle{conference}

\clearpage
\appendix
\setcounter{figure}{0}
\setcounter{table}{0}
\renewcommand{\thefigure}{\Alph{section}.\arabic{figure}}
\renewcommand{\thetable}{\Alph{section}.\arabic{table}}

\section{Additional Experimental Details and Parameter Analysis}

\subsection{Training Configuration}

We train one category-specific policy on 300 shapes from each of the furniture, containers, and daily-use groups. The joint model uses the union of all 900 shapes and is initialized by parameter averaging of the three category-specific checkpoints before another 100 epochs of PPO fine-tuning. The joint stage uses a smaller learning rate to preserve the behaviors learned by the specialist policies. As described in the main paper, advantages are normalized within each category, and the losses are first averaged per category and then macro-averaged. Tables~\ref{tab:appendix-training-settings} and~\ref{tab:appendix-model-settings} list the remaining settings and their roles.

\begin{figure}[htbp]
    \centering
    \includegraphics[width=\linewidth]{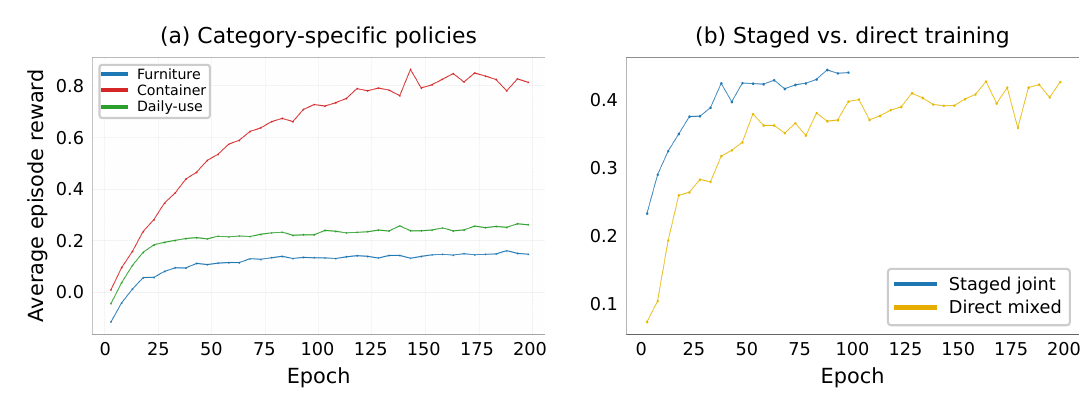}
    \caption{PPO training rewards for category-specific training and mixed-category training. Each point is averaged over a non-overlapping window of five epochs.}
    \label{fig:training-curves}
\end{figure}

\textbf{Staged versus direct mixed-category training.}
Figure~\ref{fig:training-curves} compares the optimization trajectories: joint fine-tuning from specialized policies reaches a reward near 0.44 within 100 epochs, whereas direct training on all 900 shapes reaches 0.42 after 200 epochs. Table~\ref{tab:ablation-direct-staged} likewise shows that staged training raises average mIoU from 0.5308 to 0.5460 on PartNet and from 0.4236 to 0.4476 on 3DCoMPaT++. Direct training remains applicable without category-specific initialization.

\begin{table}[htbp]
\caption{Category-level mIoU under direct mixed-category and staged joint training. Bold denotes the preferred result.}
\label{tab:ablation-direct-staged}
\centering
\small
\setlength{\tabcolsep}{5.3pt}
\begin{tabular}{lcccccccc}
\toprule
& \multicolumn{4}{c}{PartNet} & \multicolumn{4}{c}{3DCoMPaT++} \\
\cmidrule(lr){2-5}\cmidrule(lr){6-9}
Policy & Furn. & Cont. & Daily & Avg. & Furn. & Cont. & Daily & Avg. \\
\midrule
Direct mixed & 0.4186 & 0.6314 & 0.5263 & 0.5308 & 0.3558 & 0.4733 & 0.4477 & 0.4236 \\
Staged joint & \best{0.4247} & \best{0.6456} & \best{0.5432} & \best{0.5460} & \best{0.3703} & \best{0.5149} & \best{0.4607} & \best{0.4476} \\
\bottomrule
\end{tabular}
\end{table}

\begin{table}[htbp]
\caption{Data, schedule, and PPO settings for category-specific and joint training.}
\label{tab:appendix-training-settings}
\centering
\scriptsize
\setlength{\tabcolsep}{3.5pt}
\renewcommand{\arraystretch}{1.12}
\begin{tabular}{p{0.24\textwidth}ccp{0.40\textwidth}}
\toprule
Parameter & Category-specific & Joint & Role \\
\midrule
\path{num_samples} & 300 & 900 & Number of PartNet training shapes. The joint set is the union of the three category groups. \\
\path{train_epochs} & 200 & 100 & Number of outer training epochs. Each epoch collects fresh trajectories before PPO updates. \\
\path{lr} & $1\!\times\!10^{-4}$ & $5\!\times\!10^{-5}$ & Adam learning rate. The lower joint rate reduces changes to the averaged specialist initialization. \\
\path{rollout_episodes_per_epoch} & 304 & 304 & Number of complete merge-and-stop episodes collected in each outer epoch. \\
\path{rollout_workers} & 16 & 16 & Parallel workers used to collect trajectories. This changes throughput, not the objective. \\
\path{ppo_epochs} & 4 & 4 & Number of optimization passes over each collected rollout buffer. \\
\path{minibatch_size} & 128 & 128 & Number of rollout transitions in each PPO update minibatch. \\
\path{policy_temperature} & 1.0 & 1.0 & Temperature applied to masked action logits during sampling; 1.0 leaves their scale unchanged. \\
\path{seed} & 42 & 42 & Random seed for data sampling, rollouts, and model initialization. \\
\path{heartbeat_sec} & 30 & 30 & Progress-reporting interval used by MPS preprocessing; it does not affect learning. \\
\bottomrule
\end{tabular}
\end{table}

\begin{table}[htbp]
\caption{Representation, supervision, reward, and output settings shared by both training stages.}
\label{tab:appendix-model-settings}
\centering
\scriptsize
\setlength{\tabcolsep}{3.5pt}
\renewcommand{\arraystretch}{1.12}
\begin{tabular}{p{0.25\textwidth}ccp{0.39\textwidth}}
\toprule
Parameter & Category-specific & Joint & Role \\
\midrule
\path{bootstrap_source} & MPS & MPS & Uses the MPS decomposition as the initial over-segmented SQ proxy. \\
\path{point_sample_size} & 512 & 512 & Number of mesh-surface points sampled whenever the geometry observation is encoded. \\
\path{merge_neighbor_k} & 4 & 4 & Restricts each group to merge candidates among its four nearest groups. \\
\path{sq_label_method} & coverage & coverage & Assigns training-time SQ labels using geometric coverage rather than point-wise majority alone. \\
\path{final_mesh_mapping_mode} & atom & atom & Maps grouped SQs by atom voting, SQ volumetric rescue, and a final nearest-surface fallback. \\
\path{aux_part_loss_w} & 0.5 & 0.5 & Weight for the critic's auxiliary prediction of terminal part mIoU. \\
\path{aux_group_loss_w} & 0.25 & 0.25 & Weight for the critic's auxiliary prediction of terminal normalized group-count error. \\
\path{aux_target_count_loss_w} & 0.1 & 0.1 & Weight for predicting the target group-count ratio from actor-visible geometry during training. \\
\path{group_excess_penalty_w} & 0.02 & 0.02 & Dense reward weight for reducing the normalized difference between predicted and target group counts. \\
\path{stop_reward_bonus} & 0.005 & 0.005 & Base reward $b_{\mathrm{stop}}$ for an explicit stop. \\
\path{stop_group_penalty_w} & 0.1 & 0.1 & Penalizes stopping while a group-count mismatch remains. \\
\bottomrule
\end{tabular}
\end{table}

Unless stated otherwise, PPO uses discount factor $\gamma=0.99$, GAE parameter $\lambda=0.95$, clipping radius $\epsilon=0.2$, entropy coefficient $0.01$, and value-loss coefficient $0.5$. All parameter studies below use the same 300 PartNet furniture shapes and change only the stated setting.

The default maximum horizon is 200 action steps per episode, with earlier termination when the policy selects \texttt{stop}; joint training inherits the largest horizon among its source configurations. We clip the global gradient norm to 1.0. Training actions are sampled from the masked policy, whereas evaluation selects the highest-logit valid action. Point sampling remains stochastic during evaluation.

\subsection{Parameter Ablations}

\begin{figure}[htbp]
    \centering
    \includegraphics[width=0.96\linewidth]{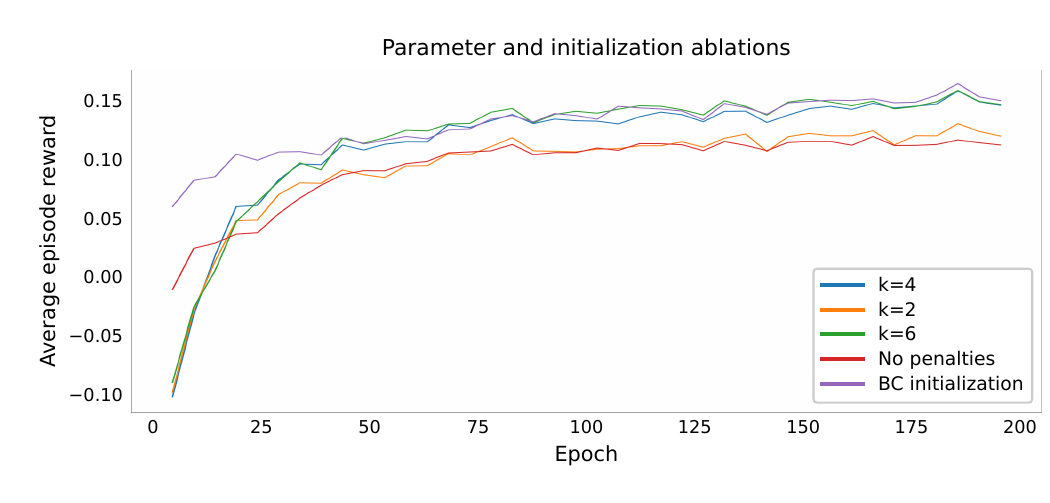}
    \caption{PPO reward curves on the 300-shape furniture training set. Curves are averaged over every five epochs. The default cold-start PPO setting uses $\kappa=4$ and both group-count penalties.}
    \label{fig:appendix-parameter-ablation}
\end{figure}

Figure~\ref{fig:appendix-parameter-ablation} reports the five-epoch mean training reward for the default setting and four variants. The default uses $\kappa=4$, both group-count shaping terms, and PPO from a random actor initialization.

\textbf{Candidate neighborhood size.}
The parameter \texttt{merge\_neighbor\_k} controls how many geometrically nearest groups are considered for each current group. With $\kappa=2$, the candidate set is often too narrow to include a useful merge, and the reward curve remains clearly below the default. Increasing the neighborhood to $\kappa=6$ produces a curve that is nearly identical to $\kappa=4$, indicating little benefit from the additional candidates. However, the larger candidate set increases the masked action space and the cost of evaluating merge features. We therefore use $\kappa=4$ as the best balance between candidate coverage and training efficiency.

\textbf{Group-count reward shaping.}
Let $G_t$ be the number of current groups and $K$ the ground-truth part count. As in the main method, their normalized mismatch is $E_t=|G_t-K|/\max(K,1)$, which treats over- and under-segmentation symmetrically. The dense term weighted by \texttt{group\_excess\_penalty\_w} rewards a transition when it reduces this mismatch, while \texttt{stop\_group\_penalty\_w} penalizes stopping with a nonzero mismatch. Setting both weights to zero removes the positive shaping signal produced by eliminating redundant groups. PPO must then assign credit mainly through changes in partition overlap, making useful merge sequences harder to learn. Accordingly, the \texttt{no\_penalties} curve converges more slowly and remains below the default, while its mIoU also decreases on both test sets in Table~\ref{tab:appendix-parameter-metrics}. For a direct measure of grouping granularity, we additionally report the mean absolute terminal count error $|G_T-K|$. Removing the penalties increases this error from $4.1500$ to $9.9667$ on PartNet and from $5.9167$ to $11.0833$ on 3DCoMPaT++. Since the SQ workspace is intentionally over-segmented and the policy progressively merges groups, the substantially lower error of the default setting shows that the count-aware rewards effectively control residual over-segmentation.

\textbf{Behavior-cloning initialization.}
We also tested whether a greedy expert could provide a useful warm start for the large merge action space. The expert scores each legal merge by its immediate post-merge mIoU, chooses the best one, and stops when no candidate gives a positive gain. We apply behavior cloning to expert trajectories from a subset of the training shapes until the cloning loss falls below 0.1, and then continue with the same PPO training used by the default model. This setting differs from the Greedy-BC baseline in the main paper, which is evaluated without subsequent PPO training. As Figure~\ref{fig:appendix-parameter-ablation} shows, the warm-started policy begins with a much higher reward, but its advantage gradually disappears and its late-stage curve overlaps the cold-start PPO curve. The expert initialization narrows early exploration without improving the final policy, adds a separate training stage, and gives lower test mIoU in Table~\ref{tab:appendix-parameter-metrics}. We therefore use cold-start PPO in RePart. Behavior cloning may still be useful for more complex tasks with much larger action spaces or rewards that are difficult to discover through unguided exploration.

\begin{table}[htbp]
\caption{Furniture results for reward-shaping and behavior-cloning variants. Count error is the mean absolute difference between the terminal group count and the ground-truth part count; lower is better. All models use the same 300 PartNet furniture training shapes.}
\label{tab:appendix-parameter-metrics}
\centering
\small
\setlength{\tabcolsep}{6pt}
\begin{tabular}{lcccc}
\toprule
& \multicolumn{2}{c}{mIoU$\uparrow$} & \multicolumn{2}{c}{$|G_T-K|\downarrow$} \\
\cmidrule(lr){2-3}\cmidrule(lr){4-5}
Training setting & PartNet & 3DCoMPaT++ & PartNet & 3DCoMPaT++ \\
\midrule
Cold-start PPO (default) & \best{0.4280} & \best{0.3699} & \best{4.1500} & \best{5.9167} \\
BC warm start $+$ PPO & 0.4165 & 0.3566 & -- & -- \\
No group-count penalties & 0.4130 & 0.3558 & 9.9667 & 11.0833 \\
\bottomrule
\end{tabular}
\end{table}

\subsection{Sensitivity to Random Seeds}

We repeat the category-specific furniture experiment with five random seeds, using 300 PartNet furniture shapes for training in each run. Table~\ref{tab:appendix-random-seeds} reports the results on the PartNet furniture test set. The final row gives the mean and sample variance across the five runs, with variance computed using a denominator of $5-1$. The mIoU ranges from 0.4247 to 0.4311, with a mean of 0.4271, indicating limited sensitivity to the random seed in this setting. The other partition metrics show some variation, particularly RI, but the results remain broadly consistent across runs rather than depending on a single favorable seed.

\begin{table}[htbp]
\caption{Random-seed sensitivity of RePart-S trained on 300 PartNet furniture shapes. The final row reports the mean, with sample variance in parentheses.}
\label{tab:appendix-random-seeds}
\centering
\small
\setlength{\tabcolsep}{10pt}
\begin{tabular}{lcccc}
\toprule
Seed & mIoU$\uparrow$ & RI$\uparrow$ & VoI$\downarrow$ & SC$\uparrow$ \\
\midrule
42    & 0.4280 & 0.7804 & 1.2997 & 0.5751 \\
91161 & 0.4266 & 0.7669 & 1.3310 & 0.5649 \\
926   & 0.4311 & 0.7967 & 1.3198 & 0.5794 \\
339   & 0.4247 & 0.7240 & 1.3500 & 0.5496 \\
32746 & 0.4253 & 0.7551 & 1.3148 & 0.5646 \\
\midrule
\shortstack[l]{Mean\\(variance)} & \shortstack{0.4271\\($6.513\times10^{-6}$)} & \shortstack{0.7646\\($7.560\times10^{-4}$)} & \shortstack{1.3231\\($3.533\times10^{-4}$)} & \shortstack{0.5667\\($1.330\times10^{-4}$)} \\
\bottomrule
\end{tabular}
\end{table}

\section{State Representation and Feature Ablations}

\subsection{State Representation}

At each decision step, the actor receives only quantities computed from the input geometry and current grouping in the Composable Primitive Workspace, which provides \emph{Structural Abstraction}. Table~\ref{tab:appendix-state-features} gives the exact organization of these inputs. Here $P=|X_s|$ is the number of points sampled for one actor observation, while $N=|X|$ denotes the full point set in the main method. We report related dimensions as feature groups instead of listing every scalar separately, while retaining the component order needed to reproduce the state. Distances and spatial extents are normalized by the scene radius unless stated otherwise; ratios are normalized by the relevant point, primitive, group, or step count. SQ parameters remain fixed after MPS initialization, whereas grouping-dependent descriptors are recomputed after every merge.

\begin{table}[htbp]
\caption{Geometry-only state used by the RePart actor. Dimensions use $M$ SQ slots, $P$ sampled points, and $C$ candidate pairs. Subgroups follow their order in the implementation.}
\label{tab:appendix-state-features}
\centering
\scriptsize
\setlength{\tabcolsep}{3pt}
\begin{tabular}{p{0.17\linewidth}c p{0.47\linewidth}p{0.25\linewidth}}
\toprule
Feature group & Dim. & Components & Update and scaling \\
\midrule
Point token & $P\!\times\!7$ & Raw xyz, centered and radius-normalized xyz, and normalized radial distance. & Points are resampled per state; center and radius are recomputed by the encoder. \\
SQ token & $M\!\times\!11$ & Two exponents, three axis lengths, three Euler angles, and three translations. & Fixed after MPS; inactive slots are zero padded and masked. \\
Group context & $M\!\times\!7$ & Nearest-SQ distance, group-size ratio, group diagonal, group-center radius, group support, candidate degree, and nearest-group AABB gap. & Recomputed after each merge from all members of the current group. \\
Assignment statistics & $M\!\times\!4$ & Normalized assignment entropy, top-1 confidence, low-margin boundary ratio, and top-1--top-2 margin. & Recomputed from soft point-to-SQ assignments. \\
Shape and fit & $M\!\times\!6$ & PCA linearity, planarity, and scattering; SQ--PCA axis alignment; mean and 90th-percentile surface residuals. & Recomputed from assigned points; residuals are divided by scene radius. \\
Global progress & $4$ & Step progress, primitive-slot occupancy, group-slot occupancy, and group count relative to initialization. & Recomputed after each action from normalized counts. \\
Global group state & $4$ & Singleton fraction, mean group-size ratio, valid-merge density, and mean candidate degree. & Recomputed from the current partition and action mask. \\
Pair compatibility & $C\!\times\!6$ & Validity, representative-center distance, representative AABB gap, scale similarity, volume ratio, and orientation similarity. & Computed for the current $\kappa$-nearest candidate pairs. \\
Pair union geometry & $C\!\times\!8$ & Combined group size, group-center distance, AABB IoU, support-ratio similarity, diagonal ratio, group AABB gap, union-volume expansion, and diagonal growth. & Uses full-group bounds and support; distance terms use scene radius. \\
Contact and orientation & $C\!\times\!3$ & AABB contact ratio, group PCA-axis alignment, and mean-normal alignment. & Recomputed for each current candidate pair. \\
\bottomrule
\end{tabular}
\end{table}

The stored point state is $P\!\times\!3$; the encoder constructs the seven-dimensional point token in Table~\ref{tab:appendix-state-features} before applying a shared MLP and concatenating max and mean pooled features. SQ parameters and the 17-dimensional group descriptors are encoded per token and pooled with \path{sq_mask}. Their pooled features and the eight global descriptors form the actor context. The stop head scores this context directly. For a merge, the actor additionally uses the two canonically ordered tokens, their absolute difference, and the 17-dimensional pair descriptor. The routing tensors---\path{action_mask} $(A)$, \path{merge_pair_indices} $(C\!\times\!2)$, \path{merge_pair_action_ids} $(C)$, and \path{merge_pair_mask} $(C)$---only map valid candidates to action logits and exclude padded or unavailable actions; they do not provide geometric evidence.

\begin{table}[htbp]
\caption{Label-derived quantities available only during training. They are excluded from the deployed actor observation.}
\label{tab:appendix-privileged-features}
\centering
\small
\setlength{\tabcolsep}{4pt}
\begin{tabular}{p{0.22\linewidth}c p{0.47\linewidth}p{0.20\linewidth}}
\toprule
State field & Dim. & Components & Use \\
\midrule
\path{privileged_prim_stats} & $M\!\times\!6$ & Point-support ratio, label purity, second-label share, unique-label ratio, normalized label entropy, and fragmentation $(1-\text{purity})$. & Critic only; recomputed after each merge. \\
\path{privileged_global_features} & $4$ & Target part-count ratio, signed and absolute group-count errors, and exact-count indicator. & Critic only; recomputed after each merge. \\
\path{target_group_count_ratio} & $1$ & Ground-truth part count divided by $M$. & Auxiliary target for the actor's geometry-based count predictor. \\
\bottomrule
\end{tabular}
\end{table}

The critic receives a detached copy of the actor context and separately encodes the SQ, primitive/group, and privileged features in Table~\ref{tab:appendix-privileged-features}. The actor predicts a target group-count ratio from geometry and provides that prediction to both action heads to judge when further merging is useful. The stored \path{target_group_count_ratio} is only its ground-truth auxiliary target and is never supplied as an actor observation. Thus, the deployed policy remains geometry-only.

\subsection{Grouped Feature Ablation}

We group related feature dimensions and remove one group at a time by setting its values to zero while preserving the network architecture and all other training settings. This avoids changing model capacity across variants. The study uses 100 PartNet furniture shapes and trains each policy for 100 epochs, with no behavior-cloning initialization. Primitive-level features are divided into group context, assignment confidence, and shape-fit statistics. Global features are divided into episode progress and current group state. Candidate-pair features are divided into pair compatibility, full-group union geometry, and contact-orientation cues.

\begin{figure}[htbp]
    \centering
    \includegraphics[width=0.96\linewidth]{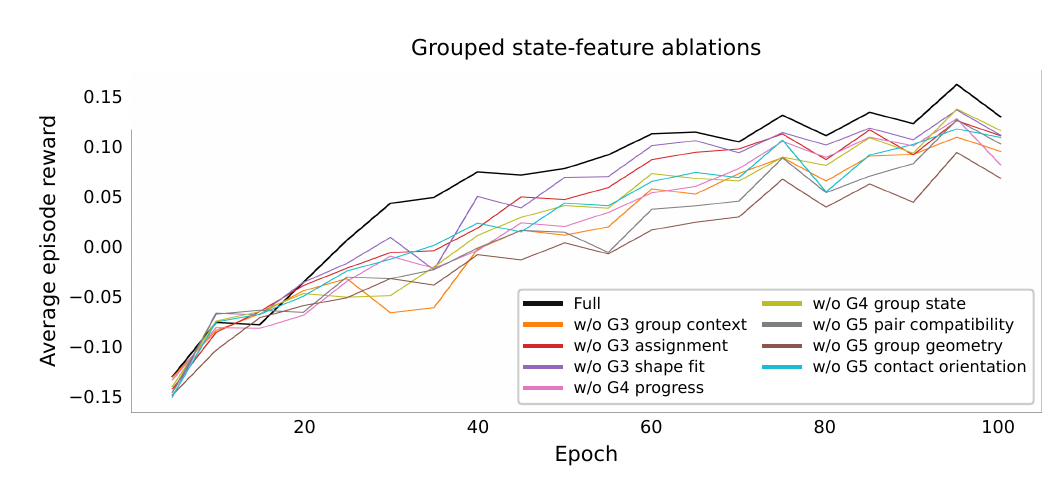}
    \caption{Grouped feature ablations on 100 PartNet furniture shapes. Curves show PPO reward averaged over every five epochs during the 100-epoch training runs. Each \emph{w/o} variant zeros only the indicated feature group.}
    \label{fig:appendix-feature-ablation}
\end{figure}

As shown in Figure~\ref{fig:appendix-feature-ablation}, the complete representation achieves the highest overall reward. Removing full-group union geometry causes the clearest degradation, because a representative primitive alone cannot describe how two multi-primitive groups would expand after merging. Pair-compatibility features are also important for rejecting nearby but structurally inconsistent candidates. Group context, assignment confidence, fitting statistics, and global progress cues have smaller individual effects, but their removals consistently remain below the complete model. These results indicate that the retained feature blocks are complementary: SQ primitives describe individual primitives, while group-level and episode-level evidence helps the policy evaluate their longer-term effect on the partition.

\subsection{Asymmetric Training Signals}

The standard asymmetric model uses label-derived information only to improve training. The per-primitive privileged statistics describe label purity, ambiguity, and fragmentation, while the privileged global features summarize target-count mismatch and terminal grouping quality. The critic combines these signals with a detached actor representation. Detachment prevents critic-side supervision from directly changing the actor encoder, while the improved value estimate can still reduce noise in PPO advantages. Separately, the ground-truth group-count ratio supervises the actor's geometry-based count predictor; at inference, the predictor operates without labels.

For the symmetric variant, we remove \path{privileged_prim_stats}, \path{privileged_global_features}, and \path{target_group_count_ratio}, disable the target-count auxiliary loss, and let the critic use only the detached actor representation. Both variants retain the same geometry-only actor, action space, and reward definition, and ground-truth labels remain available only for computing the training reward. This comparison therefore isolates the contribution of asymmetric training rather than changing the inference inputs.

\begin{figure}[htbp]
    \centering
    \includegraphics[width=0.90\linewidth]{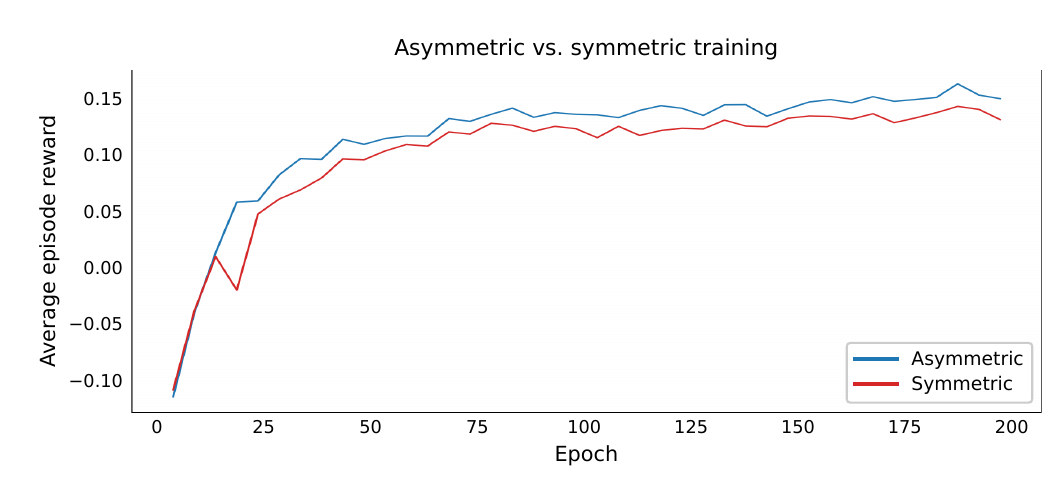}
    \caption{Asymmetric and symmetric training on 300 PartNet furniture shapes for 200 epochs. Curves show PPO reward averaged over every five epochs. The asymmetric variant uses label-derived critic features and target-count supervision only during training; both actors use geometry-only inputs.}
    \label{fig:appendix-asymmetric}
\end{figure}

Figure~\ref{fig:appendix-asymmetric} shows that the asymmetric model maintains a higher reward through most of training and finishes above the symmetric variant. Privileged purity and fragmentation statistics help the critic distinguish states that can appear geometrically similar but differ in partition quality, producing more informative advantage estimates for earlier merge decisions. Count supervision also provides a direct training signal for stopping granularity. The gain is moderate rather than dominant, which agrees with the main design: geometric state features drive the policy, while asymmetric information makes their optimization more effective without introducing label dependence at inference.

\subsection{Future Work: LLM- and Agent-Based Policy Generation}
\label{sec:appendix-agent-policy}

The Composable Primitive Workspace separates semantic organization from dense surface prediction and is not tied to a particular policy architecture. This suggests exploring pretrained large language models (LLMs) or multimodal agents as alternatives to the current PPO-based policy in Reinforcement-Learning Part Reasoning. Compact primitives and explicit group relations could provide a more manageable basis for such models than direct reasoning over dense mesh elements. Related work on language-guided shape editing illustrates the potential of combining foundation models with geometric representations \citep{ganeshan2024parsel}. In our setting, an LLM or agent might guide merge-and-stop decisions while retaining the Composable Primitive Workspace and Boundary-Aware Surface Labeling, potentially avoiding task-specific RL training.

Such an alternative may draw on broader knowledge of object composition and reduce dependence on the limited category coverage of our training set, offering a possible route toward better generalization to unfamiliar structures. However, broader prior knowledge does not guarantee geometrically consistent grouping. Reliable stopping and label-free feedback remain open questions, since our current reward and privileged critic depend on ground-truth annotations unavailable at deployment. We have not implemented or evaluated this extension; its effectiveness and computational cost relative to PPO remain topics for future work.

\section{Boundary-Aware Surface Labeling}
\label{sec:appendix-projection}

We ablate the two components of Boundary-Aware Surface Labeling that transfer terminal SQ groups back to mesh faces. All variants reuse exactly the same grouped-SQ predictions, raw meshes, SQ surface samples, and ground-truth face correspondences, so the comparison isolates Boundary-Aware Surface Labeling rather than Reinforcement-Learning Part Reasoning. We evaluate three increasingly structured variants on the daily-use, furniture, and container groups:
\begin{itemize}[leftmargin=1.5em,itemsep=0pt,topsep=2pt]
    \item \textbf{Facewise nearest SQ} assigns each face centroid independently to the group with the nearest sampled SQ surface point.
    \item \textbf{+ Curvature atoms} first over-segments the mesh along high-dihedral adjacencies. Valid proximity votes are aggregated within each atom, while rejected atoms are resolved by the common nearest-surface fallback. SQ volumetric rescue is disabled in this variant.
    \item \textbf{+ SQ volumetric evidence} is the complete Boundary-Aware Surface Labeling module. It reconsiders rejected atoms using the analytic signed-distance evidence of the grouped SQs before applying the same fallback to any remaining rejected atom.
\end{itemize}

For reproducibility, each SQ surface is sampled on a $72\times36$ parameter grid. The atom boundary threshold is the 90th percentile of adjacent-face dihedral angles, clipped to $[6^\circ,12^\circ]$, followed by one label-smoothing iteration. We use uniform vote counts, a normalized face-distance threshold of $0.04$, a maximum distant-face fraction of $0.6$, and a minimum atom size of one face. For volumetric rescue, an atom is accepted when at least half of its centroids support one group with analytic SQ signed distance no greater than $0.005$. This evidence is evaluated only for rejected atoms and therefore does not overwrite reliable proximity assignments. If both tests fail, an unrestricted nearest-surface majority vote assigns the whole atom. All partition metrics are computed after this common fallback over the same complete set of mesh faces with valid ground-truth labels; no unassigned prediction is excluded. Final coverage is therefore 100\% for every variant. We additionally report the fraction of faces assigned by fallback as a diagnostic of how often the preceding stages remain unresolved.

\begin{table}[htbp]
\caption{Ablation of Boundary-Aware Surface Labeling on PartNet. Partition metrics are evaluated on all valid faces after the common fallback. Final coverage is 100\% for every variant; FB faces denotes the percentage assigned only by fallback and does not apply to direct facewise nearest assignment. Bold denotes the preferred result.}
\label{tab:appendix-projection}
\centering
\scriptsize
\setlength{\tabcolsep}{4.5pt}
\begin{tabular}{llccccc}
\toprule
Category & Projection & mIoU$\uparrow$ & SC$\uparrow$ & VoI$\downarrow$ & Final cov. & FB faces$\downarrow$ \\
\midrule
Daily-use & Facewise nearest SQ & 0.4500 & 0.5574 & 1.2427 & 1.0000 & -- \\
Daily-use & $+$ Curvature atoms & 0.5351 & 0.6597 & 0.8608 & 1.0000 & 3.7864\% \\
Daily-use & $+$ SQ volumetric evidence & \best{0.5432} & \best{0.6598} & \best{0.8607} & 1.0000 & \best{3.1309\%} \\
\midrule
Furniture & Facewise nearest SQ & 0.3310 & 0.4645 & 1.8139 & 1.0000 & -- \\
Furniture & $+$ Curvature atoms & 0.4203 & \best{0.5536} & \best{1.3358} & 1.0000 & 6.8709\% \\
Furniture & $+$ SQ volumetric evidence & \best{0.4247} & 0.5522 & 1.3373 & 1.0000 & \best{0.9656\%} \\
\midrule
Container & Facewise nearest SQ & 0.6139 & 0.7868 & 0.5791 & 1.0000 & -- \\
Container & $+$ Curvature atoms & 0.6421 & 0.8280 & \best{0.4256} & 1.0000 & 2.9452\% \\
Container & $+$ SQ volumetric evidence & \best{0.6456} & \best{0.8282} & 0.4257 & 1.0000 & \best{2.2130\%} \\
\bottomrule
\end{tabular}
\end{table}

Table~\ref{tab:appendix-projection} evaluates the same complete face set used by the deployed output. The complete Boundary-Aware Surface Labeling module improves mIoU over facewise nearest assignment by $0.0932$, $0.0937$, and $0.0317$ on daily-use, furniture, and container objects, respectively. The corresponding SC gains are $0.1024$, $0.0877$, and $0.0414$, while VoI decreases by $0.3820$, $0.4766$, and $0.1534$. These consistent final-output gains show that regional voting suppresses fragmented labels and better aligns group transitions with mesh structure.

Curvature atoms account for most of the metric improvement, while SQ volumetric evidence mainly resolves regions that surface voting leaves uncertain. It further raises mIoU from $0.5351$ to $0.5432$ on daily-use objects, from $0.4203$ to $0.4247$ on furniture, and from $0.6421$ to $0.6456$ on containers, while SC and VoI remain nearly unchanged. More importantly, it reduces the fallback-face rate from $3.7864\%$ to $3.1309\%$ for daily-use objects, from $6.8709\%$ to $0.9656\%$ for furniture, and from $2.9452\%$ to $2.2130\%$ for containers, corresponding to relative reductions of $17.31\%$, $85.95\%$, and $24.86\%$. Thus, volumetric evidence reduces reliance on the unrestricted fallback, while the shared fallback guarantees a complete final partition for every variant.

\section{Generalization to Unseen PartNet Categories}
\label{sec:appendix-unseen}

Our PartNet training set contains 14 categories grouped as containers (Bag, Bottle, Bowl, Mug, and Vase), furniture (Chair, Door, StorageFurniture, and Table), and daily-use objects (Earphone, Faucet, Knife, Lamp, and Scissors). The main PartNet evaluation uses held-out shapes from these category groups. To study transfer beyond the training categories, we additionally test on Bed, Display, Hat, Laptop, Refrigerator, and TrashCan, which are excluded from training but share some part-level structures with the seen categories. We randomly sample ten shapes from each unseen category and use the same evaluation protocol as in the main experiments.

\begin{table}[htbp]
\caption{Part segmentation on PartNet categories excluded from training. Results are averaged over the six unseen categories. Bold and underline denote the best and second-best results, respectively.}
\label{tab:appendix-unseen}
\centering
\small
\setlength{\tabcolsep}{8pt}
\begin{tabular}{lcccc}
\toprule
Method & mIoU$\uparrow$ & RI$\uparrow$ & VoI$\downarrow$ & SC$\uparrow$ \\
\midrule
MPS & 0.2103 & 0.4460 & 1.1156 & 0.4460 \\
PartField & \best{0.4295} & \best{0.7416} & 1.1886 & \underline{0.5875} \\
P3-SAM & 0.3834 & \underline{0.7112} & \underline{1.1090} & 0.5851 \\
SAMPart3D & 0.2393 & 0.5380 & 1.3913 & 0.4574 \\
PriMAny & 0.2028 & 0.5793 & 2.7416 & 0.2379 \\
\midrule
\textbf{RePart (Ours)} & \underline{0.3992} & 0.7033 & \best{0.9327} & \best{0.5934} \\
\bottomrule
\end{tabular}
\end{table}

\begin{table}[htbp]
\caption{Category-level mIoU on PartNet categories excluded from training. Bold and underline denote the best and second-best results, respectively.}
\label{tab:appendix-unseen-category}
\centering
\scriptsize
\setlength{\tabcolsep}{4.2pt}
\begin{tabular}{lcccccc}
\toprule
Category & MPS & PartField & P3-SAM & SAMPart3D & PriMAny & RePart \\
\midrule
Bed          & 0.0789 & \best{0.2908} & 0.2313 & 0.1526 & 0.2094 & \underline{0.2661} \\
Display      & 0.1980 & \underline{0.4304} & 0.4301 & 0.2888 & 0.2000 & \best{0.4693} \\
Hat          & 0.4417 & 0.6135 & \best{0.7073} & 0.3813 & 0.1860 & \underline{0.6222} \\
Laptop       & 0.1662 & \underline{0.3291} & 0.3184 & 0.2097 & 0.2385 & \best{0.3672} \\
Refrigerator & 0.1330 & \best{0.3337} & 0.1930 & 0.1464 & 0.1289 & \underline{0.2554} \\
TrashCan     & 0.2439 & \best{0.5791} & \underline{0.4206} & 0.2530 & 0.2539 & 0.4153 \\
\bottomrule
\end{tabular}
\end{table}

\begin{figure}[htbp]
    \centering
    \includegraphics[width=\linewidth]{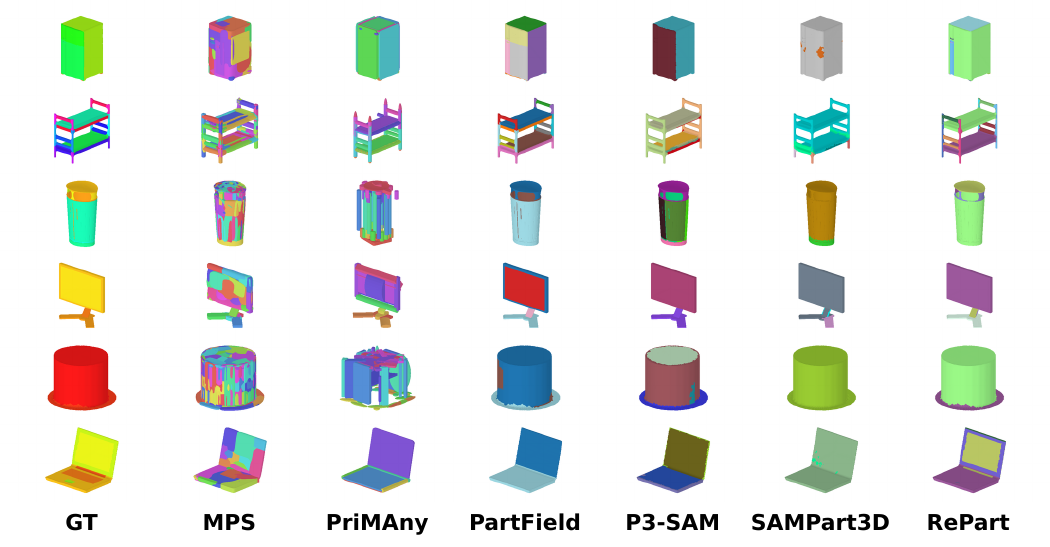}
    \caption{Qualitative comparison on PartNet categories excluded from training.}
    \label{fig:appendix-unseen-visuals}
\end{figure}

As shown in Table~\ref{tab:appendix-unseen}, RePart achieves the best VoI and SC and the second-best average mIoU, while remaining competitive in RI. Table~\ref{tab:appendix-unseen-category} further shows that it ranks first or second in mIoU on five of the six unseen categories, including the best results on Display and Laptop. The low VoI indicates that its predicted groups preserve coherent part structure rather than fragmenting objects excessively. These results and the examples in Figure~\ref{fig:appendix-unseen-visuals} show that the learned merge policy retains useful transfer to unseen categories when they contain local structures similar to those observed during training. This experiment supports generalization across related structures, rather than unrestricted open-category segmentation.

\section{Additional Qualitative Comparisons}
\label{sec:appendix-qualitative}

\begin{figure}[htbp]
    \centering
    \includegraphics[width=\linewidth,height=0.80\textheight,keepaspectratio]{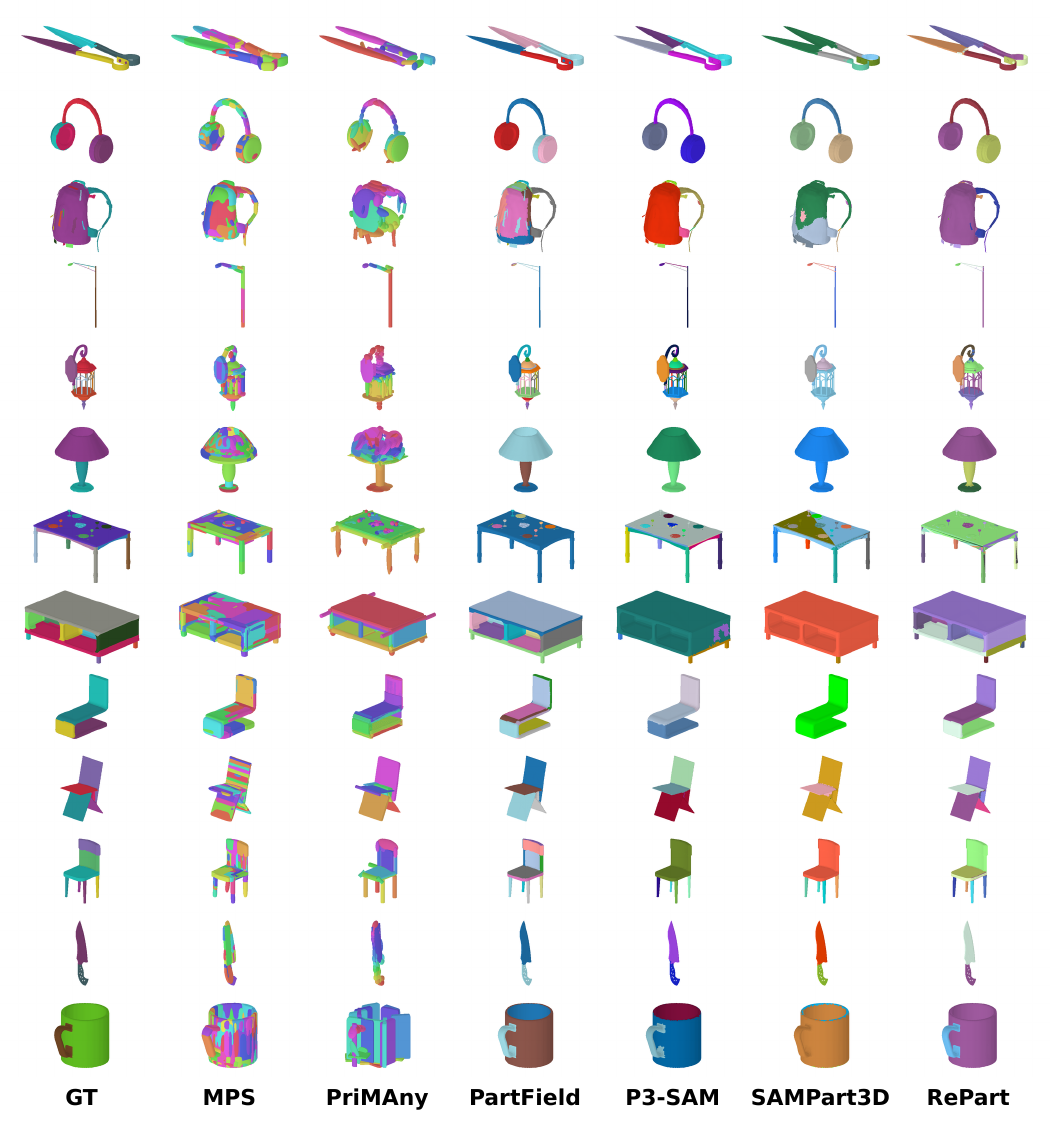}
    \caption{Additional qualitative comparisons on PartNet.}
    \label{fig:appendix-partnet-visuals}
\end{figure}

\begin{figure}[htbp]
    \centering
    \includegraphics[width=\linewidth,height=0.82\textheight,keepaspectratio]{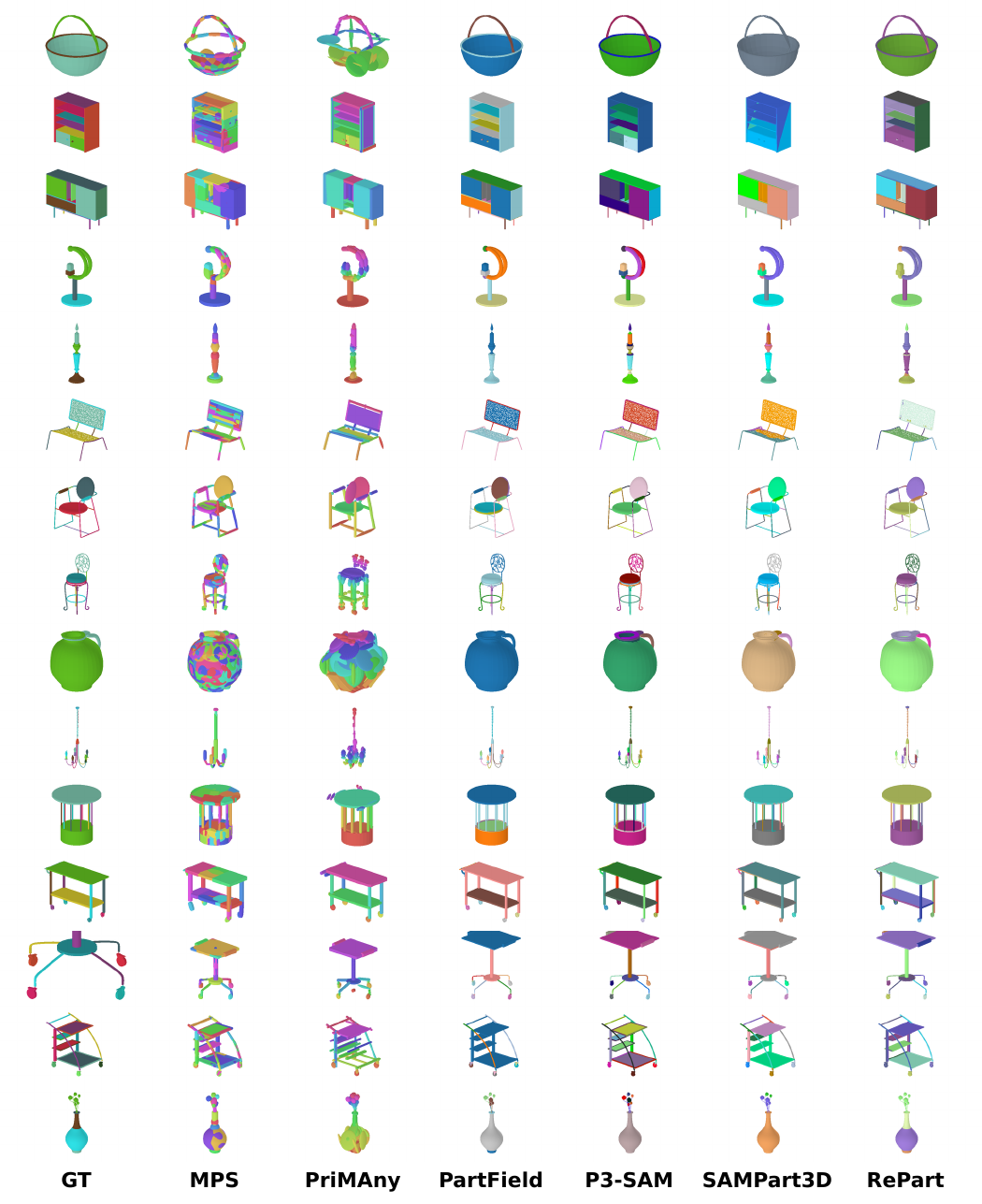}
    \caption{Additional cross-dataset qualitative comparisons on 3DCoMPaT++.}
    \label{fig:appendix-compat-visuals}
\end{figure}

Figures~\ref{fig:appendix-partnet-visuals} and~\ref{fig:appendix-compat-visuals} extend the qualitative comparison in Figure~\ref{fig:qualitative-results}. 

\clearpage

\section{Superquadric Distance and Surface Evaluation}
\label{app:sq-distance}
This section records the standard SQ functions used for point assignment and volumetric evidence in the main text. It does not introduce an additional fitting or inference stage. For $p_j=(\ve_j,\va_j,\vomega_j,\vt_j)$, let $\bar{\vx}_j=R(\vomega_j)^\top(\vx-\vt_j)$, where $R$ maps ZYX Euler angles to a rotation matrix. The inside--outside function is~\citep{barr1981superquadrics,liu2023marching}
\begin{equation}
F_j(\vx)=\left[
\left|\frac{\bar x_{j,x}}{a_{j,x}}\right|^{2/e_{j,2}}+
\left|\frac{\bar x_{j,y}}{a_{j,y}}\right|^{2/e_{j,2}}
\right]^{e_{j,2}/e_{j,1}}+
\left|\frac{\bar x_{j,z}}{a_{j,z}}\right|^{2/e_{j,1}}.
\label{eq:sq-function}
\end{equation}
The level set $F_j(\vx)=1$ defines the analytic surface sampled by $\mathcal{B}(p_j)$. Homogeneity along rays from the SQ center gives the radial signed-distance approximation~\citep{solina1990recovery,liu2023marching}
\begin{equation}
d_{\mathrm{SQ}}(\vx;p_j)=\|\bar{\vx}_j\|_2
\left(1-F_j(\vx)^{-e_{j,1}/2}\right).
\label{eq:dsq}
\end{equation}
It is positive outside the primitive and negative inside; it is a radial approximation, not an exact Euclidean signed distance. Its absolute value supplies the distance $d_{ij}$ used in point-to-primitive assignment, while its sign supplies the volumetric evidence $D_u$ used in Boundary-Aware Surface Labeling.

\end{document}